\documentclass{article} 
\usepackage{iclr2027_conference,times}

\usepackage{subcaption}
\usepackage{multirow}
\usepackage{pgfplots}\pgfplotsset{compat=1.17}
\usepgfplotslibrary{fillbetween}
\definecolor{basecol}{HTML}{2166AC}
\definecolor{sftcol}{HTML}{B8860B}
\usepackage{wrapfig}
\usepackage{pifont}
\usepackage{multirow}
\usepackage{microtype}
\usepackage{url}
\usepackage{longtable}
\usepackage{booktabs}
\usepackage[utf8]{inputenc}
\usepackage{graphicx}
\usepackage{flafter}
\usepackage{amsmath, amssymb, amsthm}
\usepackage{caption}
\usepackage{subcaption}
\usepackage{float}
\usepackage{enumitem}
\usepackage{algorithm}
\usepackage{algpseudocode}
\usepackage{hyperref}
\usepackage[table]{xcolor}
\usepackage{colortbl}
\usepackage{makecell}
\usepackage{wrapfig}
\usepackage{soul}
\usepackage{xcolor}
\usepackage{newunicodechar}
\newunicodechar{â}{-}
\definecolor{darkblue}{rgb}{0,0,0.5}
\definecolor{darkyellow}{HTML}{B8860B} 
\newcommand{\hyp}[1]{\textbf{\textcolor{darkyellow}{#1}}} 
\newcommand{\hi}[1]{\textcolor{darkyellow}{\textbf{#1}}} 
\definecolor{ourrow}{HTML}{F3E7B8}  
\definecolor{hdrband}{HTML}{EAD98A}  
\definecolor{highlightgreen}{RGB}{144,238,144}
\definecolor{highlightblue}{RGB}{230,240,250}

\hypersetup{colorlinks=true, citecolor=darkyellow, linkcolor=darkyellow, urlcolor=darkyellow}

\usepackage{array}
\definecolor{ourscol}{HTML}{F7EFCB}  
\definecolor{headgray}{RGB}{223,223,223}  
\definecolor{winblue}{RGB}{28,78,140}     
\definecolor{wingreen}{RGB}{27,94,32}     
\newcolumntype{O}{>{\columncolor{ourscol}}c} 
\newcolumntype{Y}{>{\columncolor{gray!8}}c}  
\newcommand{\wc}[1]{\textcolor{darkyellow}{\textbf{#1}}} 
\newcommand{\hdr}{\rowcolor{hdrband}}         
\newcommand{\grp}{\rowcolor{hdrband!45}}      
\newcommand{\winrow}{\rowcolor{ourrow}}       

\algrenewcommand\algorithmicindent{1.0em}

\usepackage[most]{tcolorbox}
\usepackage[fontsize=9.5pt]{fontsize} 
\usepackage[margin=1in]{geometry} 
\usepackage{sectsty} 
\sectionfont{\color{darkyellow}}
\subsectionfont{\color{darkyellow}}
\subsubsectionfont{\color{darkyellow}}
\newcommand{\blfootnote}[1]{%
  \begingroup\renewcommand{\thefootnote}{}\footnotetext{#1}%
  \addtocounter{footnote}{-1}\endgroup}
\definecolor{darkgreen}{HTML}{1B5E20}
\definecolor{promptframe}{HTML}{37474F}

\newtcolorbox{descbox}[1]{
  enhanced, breakable, rounded corners,
  colback=yellow!8!white, colframe=darkyellow,
  coltitle=white, title={\textbf{#1}}, fonttitle=\bfseries,
  left=6pt, right=6pt, top=4pt, bottom=4pt}

\newtcolorbox{remarkbox}[1][Remark]{
  enhanced, breakable, rounded corners,
  colback=orange!5!white, colframe=orange!72!black, boxrule=0.6pt,
  title={\textbf{#1}}, fonttitle=\bfseries\small, fontupper=\small,
  left=6pt, right=6pt, top=4pt, bottom=4pt, before skip=8pt, after skip=8pt}
\newtcolorbox{recipebox}[1][Overview]{
  enhanced, breakable, sharp corners,
  colback=yellow!8!white, colframe=darkyellow, boxrule=0.5pt,
  title={\textbf{#1}}, fonttitle=\bfseries\footnotesize, fontupper=\footnotesize,
  left=6pt, right=6pt, top=3pt, bottom=3pt, before skip=6pt, after skip=6pt}
\definecolor{promptbg}{HTML}{F4F6F7}   
\newtcolorbox{promptbox}[1]{
  enhanced, breakable, rounded corners,
  colback=promptbg, colframe=darkyellow, boxrule=0.7pt,
  colbacktitle=darkyellow, coltitle=white, attach boxed title to top left={yshift=-2pt,xshift=8pt},
  boxed title style={rounded corners, boxrule=0pt},
  title={\textbf{#1}}, fonttitle=\bfseries\small, fontupper=\small,
  left=6pt, right=6pt, top=6pt, bottom=4pt, before skip=8pt, after skip=8pt}

\title{Teaching Agents to Code Reliably}

\author{Muhammad Ahmed Mohsin\thanks{Work done during an internship at AWS AI Labs.} \\
Stanford University \\
\texttt{muahmed@stanford.edu} \\
\And
Myeongsoo Kim, Kangrui Ruan, Shweta Garg, Varun Kumar, Murali Krishna Ramanathan \\
AWS AI Labs \\
\texttt{\{mysoo, kangruir, shwegarg, kuvrun, mkraman\}@amazon.com}
}

\iclrfinalcopy 
\begin{document}
\pagestyle{plain}\thispagestyle{plain} 

\definecolor{cardbg}{HTML}{FBF5EA}      
\newcommand{\amazonlogo}{%
  \IfFileExists{logo.png}%
    {\includegraphics[height=1.35\baselineskip]{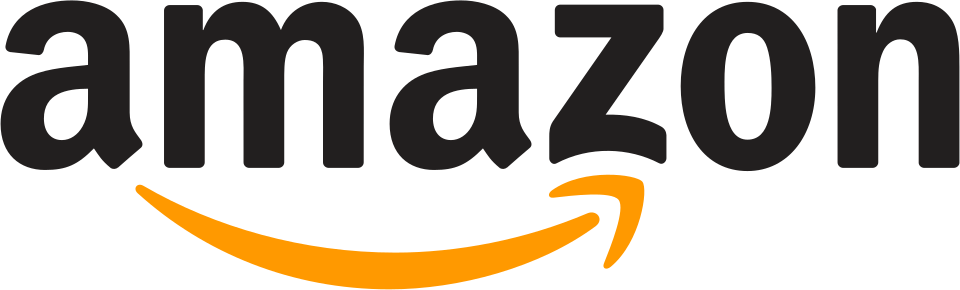}}%
    {\raisebox{1pt}{\large\bfseries\textcolor{darkyellow}{amazon}}}%
}

\begin{tcolorbox}[enhanced, breakable, halign=justify,
  colback=cardbg, colframe=cardbg, boxrule=0pt, arc=4mm,
  left=8mm, right=8mm, top=7mm, bottom=8mm, drop shadow=black!10]
{\raggedright\LARGE\bfseries Teaching Agents to Code Reliably\par}
\vspace{4mm}
{\raggedright\footnotesize\bfseries Muhammad Ahmed Mohsin\textsuperscript{1,}\textsuperscript{$*$}, Myeongsoo Kim\textsuperscript{2}, Kangrui Ruan\textsuperscript{2}, Shweta Garg\textsuperscript{2}, Varun Kumar\textsuperscript{2}, Murali Krishna Ramanathan\textsuperscript{2}\par}
\vspace{2mm}
{\centering\footnotesize\bfseries \textsuperscript{1}Stanford University \qquad \textsuperscript{2}AWS AI Labs\par}
\vspace{1mm}
{\centering\footnotesize\ttfamily \{cmohsinm,\,mysoo,\,kangruir,\,shwegarg,\,kuvrun,\,mkraman\}@amazon.com\par}
\vspace{4mm}
{\color{black!15}\hrule height 0.6pt}
\vspace{3.5mm}
{\fontsize{8.5}{11}\selectfont
\noindent\textbf{Abstract.}\ Autonomous coding agents solve repository issues by reading code, running commands, editing files, and submitting patches. Extra inference-time compute yields gains only when it produces a useful repair and supplies reliable evidence for choosing one. \hyp{Three behaviors decide both, and we argue they are teachable rather than byproducts of scale, so a policy can carry them instead of a scaffold.} Location diversity remains narrow, since attempts return to the same site and extra samples add no coverage. Edit diversity is left unexploited, since methodologies that differ resolve complementary issues no single run reaches. Verification misleads, since a test the agent writes for its own patch accepts many incorrect ones. Directing search by execution feedback and scoring each patch against its own reverted tree resolves $52.8\%$ of SWE-bench Verified using $48.1\%$ of the agent-steps an eight-sample baseline spends. Training moves these behaviors into the policy. On the $270$ issues held out from SFT and RL training, weighted supervised fine-tuning raises pass@$1$ from $31.9\%$ to $35.2\%$ and pass@$8$ from $46.7\%$ to $51.1\%$. A reinforcement objective then trains the verifier against gold-labeled repairs and incorrect variants, crediting the assertions that detect them. It raises pass@$1$ to $43.0\%$ and pass@$8$ to $60.7\%$, lifts verifier precision from $26.8\%$ to $41.7\%$, and more than halves false acceptance. Resolution improves on two of three out-of-distribution suites and verifier precision on all three, and the gains hold at $7$B, $14$B, and $30$B against published coder baselines.
\par}
\vspace{6mm}
\par\noindent\textbf{Date:}~18 September 2026\par
\noindent\textbf{Correspondence:}~\texttt{muahmed@stanford.edu}\hfill\amazonlogo
\end{tcolorbox}
\blfootnote{\textsuperscript{$*$}Work done during an internship at AWS AI Labs.}
\vspace{2mm}

\section{Introduction}
\label{sec:introduction}
Autonomous coding agents resolve repository issues through code inspection, editing, and test execution \citep{jimenez2024swebench,yang2024sweagent}. Repeated attempts and verification can improve resolution by increasing inference-time computation \citep{brown2024monkeys,snell2024scaling,han2026swetrace}. Their benefit depends on whether additional attempts produce correct repairs and whether the agent can identify them. Attempts can revisit the same edit location or repeat an incorrect repair. A test generated by the agent can share the patch's mistaken interpretation of the issue and accept incorrect behavior \citep{ahmed2025testoverfitting}. More candidates do not guarantee a correct selected patch.

We distinguish candidate coverage from selection quality. Coverage is the fraction of issues for which the candidate set contains at least one correct repair. Selection quality is the fraction of those issues for which the agent returns a correct candidate. Oracle pass@$8$ measures coverage across eight attempts, whereas the accuracy of the selected patch depends on both quantities. Verification also guides the search by determining whether the agent revises a patch, explores another location, or stops. The verifier therefore affects both the candidates generated and the patch returned.

We hypothesize that training can improve repair search and test discrimination under a fixed backbone and scaffold. Our diagnostic study examines edit locations, repair approaches, and verification decisions on SWE-bench Verified. Repeated self-verification produces few changes to the submitted patches, while different problem-solving preambles produce complementary repairs. Proactive localization visits more locations than reactive search but resolves fewer issues (Section~\ref{sec:tts}). These observations motivate training the agent to use execution feedback when choosing its next edit and to evaluate tests by their ability to detect incorrect repairs. Execution provides evidence beyond the agent's judgment of its own output \citep{huang2024selfcorrect}.

Our approach combines a test-time diagnostic with two training stages. The diagnostic pairs exploration of edit locations and repair approaches with differential verification, which runs a reproduction on the patched and reverted repository. Weighted supervised fine-tuning uses the resulting trajectories, with weights determined by repair-quality tier, exploration score, and agent-step count. Reinforcement learning then evaluates generated tests against oracle-rejected patch variants and sibling repairs. On the $270$ SWE-bench Verified issues held out from SFT and RL training, the two training stages raise greedy pass@$1$ from $31.9\%$ to $43.0\%$ (Table~\ref{tab:rl-main}). Across eight attempts, oracle coverage rises from $46.7\%$ to $60.7\%$ and selected-patch accuracy from $33.5\%$ to $50.5\%$, with selection efficiency increasing from $0.72$ to $0.83$ (Table~\ref{tab:e6-selector}). These measurements separate improvements in candidate coverage from improvements in selection. On the external benchmarks, RL raises pass@$1$ over SFT from $15.7\%$ to $20.2\%$ on SWE-bench Pro and from $24.2\%$ to $31.8\%$ on SWE-PolyBench, while DeepSWE full resolution remains zero at this scale, though its partial-credit measures rise (Table~\ref{tab:rl-benchmarks}).

\begin{figure*}[t!]
    \centering
    \includegraphics[width=0.8\linewidth]{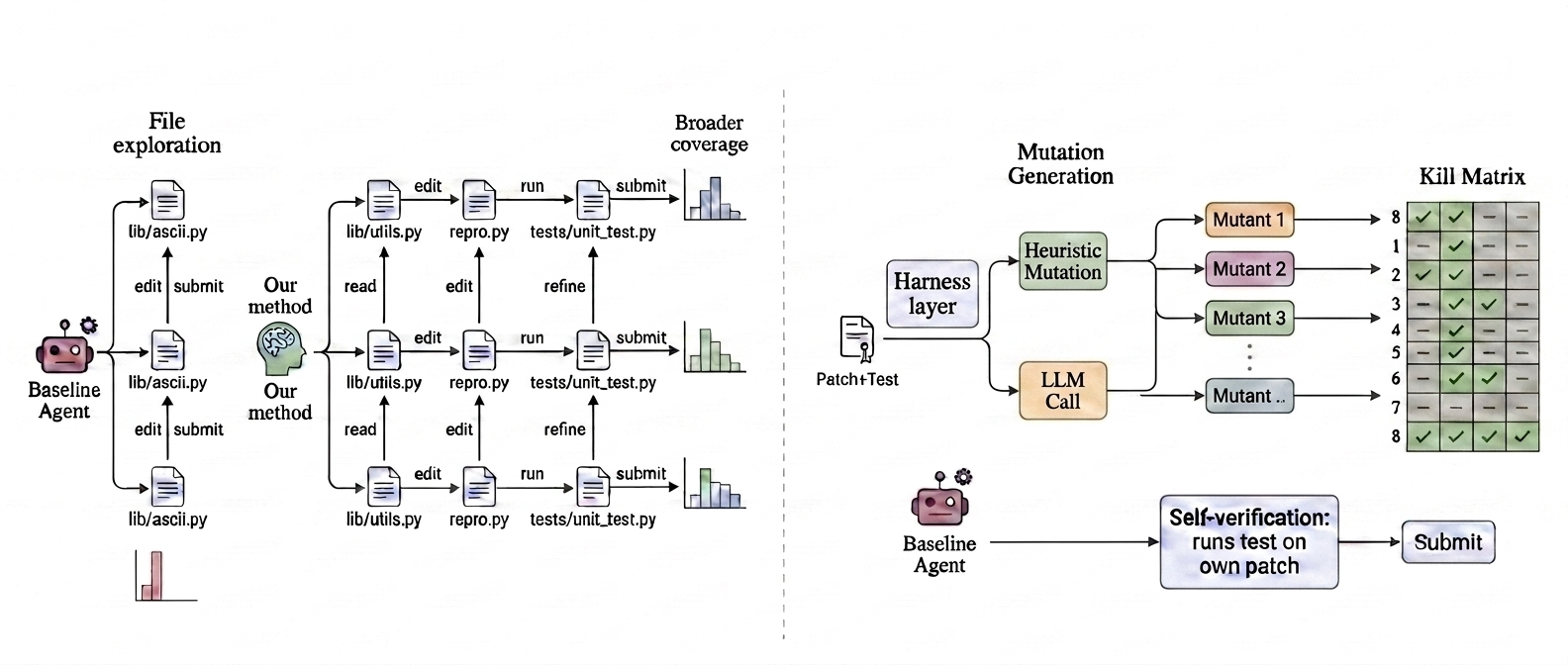}
    \caption{\textbf{Two behaviors, one reward.}
  \emph{Left, exploration.} The baseline resubmits the same site (\texttt{lib/ascii.py}), so coverage
  stays narrow, while parallel chains that read first each reach the true defect
  (\texttt{lib/utils.py}), write a reproduction, and submit. \emph{Right, verification.} The harness
  creates incorrect patch variants and records which assertions detect each variant. The gate opens only when the test kills
  every variant, so the reward pays for discrimination.}
    \label{fig:methodology}
\end{figure*}

\section{Methodology}
\label{sec:methodology}
This section describes the test-time search procedures, weighted trajectory supervision, and reinforcement learning objective. Figure~\ref{fig:methodology} illustrates exploration and verification within one repair trajectory.

\subsection{Test-Time Scaling Methods on Coding Agents}
\label{sec:tts}

The diagnostic uses all $500$ SWE-bench Verified issues to analyze repair and verification failures. We select $230$ issues from this population for training dataset curation and hold the remaining $270$ out of SFT and RL training. All search procedures run on the same backbone and scaffold. Table~\ref{tab:efficiency-all} reports resolution, agent-steps, attempts per instance, and diversity. Attempt limits and stopping rules differ by procedure, so each comparison reflects an accuracy and cost trade-off.

We compare single attempts, repeated self-verification, eight repair preambles, proactive localization, and reactive bans on failed edits or locations. The preambles reach an oracle union of $305/500$ issues, while repeated self-verification leaves $78$ to $85\%$ of audit passes without a code change. Appendix~\ref{app:search-details} describes the procedures and scoring rules.

Differential-verification search pairs up to eight locations and repair approaches. It runs each reproduction on the patched and reverted trees, checks repository tests and changed-line execution, and uses the resulting score to stop or prune search. A reproduction that passes on both trees receives zero credit. Search uses repository execution rather than hidden grader feedback. It resolves $52.8\%$ of issues against $46.8\%$ for the eight-sample control, using $48.1\%$ of its agent-steps (Table~\ref{tab:efficiency-all}). Appendix~\ref{tab:e9e-dvs} reports component ablations.

\begin{table}[t]
\centering
\small
\setlength{\tabcolsep}{4pt}
\renewcommand{\arraystretch}{0.95}
\resizebox{\textwidth}{!}{%
\begin{tabular}{l cc cccc cc}
\toprule
\rowcolor{hdrband}
& \multicolumn{2}{c}{\textbf{Accuracy}} & \multicolumn{4}{c}{\textbf{Compute}}
& \multicolumn{2}{c}{\textbf{Diversity}}\\
\rowcolor{hdrband}
\textbf{Experiment} & \textbf{/500} & \textbf{rate}
& \textbf{total steps} & \textbf{att./inst} & \textbf{early-stop} & \textbf{steps/solve}
& \textbf{loc.} & \textbf{edit} \\
\midrule
\rowcolor{gray!8} Self-Verification            & 184 & 36.8\% & 189{,}371 & 3.99 & 0 / 500  & 1{,}029 & 1.10 & 1.31 \\
\rowcolor{gray!8} Single-Axis Diagnosis        & 189 & 37.8\% & \hi{41{,}332} & 1.00 & $-$ & \hi{219} & $-$ & $-$ \\
\rowcolor{gray!8} Proactive Location Diversity & 210 & 42.0\% & 138{,}424 & 4.00 & 0 / 500  & 659 & \hi{2.46} & \hi{3.26} \\
\rowcolor{gray!16}\quad\emph{Blind Retry (baseline)} & 213 & 42.6\% & 139{,}632 & $-$ & $-$ & 656 & $-$ & $-$ \\
\rowcolor{gray!8} Reactive Edit Diversity      & 217 & 43.4\% & 131{,}417 & 3.67 & 65 / 500 & 606 & 1.84 & 2.79 \\
\rowcolor{gray!8} Reactive Location Diversity  & 225 & 45.0\% & 131{,}680 & 3.66 & 66 / 500 & 585 & 1.86 & 2.73 \\
\rowcolor{gray!16}\quad\emph{Independent Retry (baseline)} & 221 & 44.2\% & 130{,}564 & $-$ & $-$ & 591 & $-$ & $-$ \\
\rowcolor{gray!16}\emph{Sample Scaling} & 234 & 46.8\% & 275{,}943 & 8.00 & 0 / 500 & 1{,}179 & $\sim$2.0 & $-$ \\
\rowcolor{ourrow}\textbf{Differential Verification} & \hi{264} & \hi{52.8\%} & 132{,}697 & 3.12 & \hi{102 / 500} & \textbf{503} & 1.75 & $-$ \\
\bottomrule
\end{tabular}%
}
\caption{Efficiency across the test-time procedures on the $500$ issues, all-attempts basis, so
steps/solve equals total steps over bugs resolved. Dark yellow marks the best value in a column. The
highlighted row is our combined procedure, which is strongest on accuracy at near the lowest cost.
Single-axis diagnosis is the one-attempt reference, cheapest overall though not the lowest resolution.
Methodology diversity is reported in the text and not here, because its $61.0\%$ is an oracle union over
eight preambles rather than a rate any single deployable procedure reaches. Blind Retry,
Independent Retry, and Sample Scaling are baselines in italics. A dash marks not applicable.}
\label{tab:efficiency-all}
\end{table}

\subsection{Supervised Fine-Tuning on Weighted Agent Trajectories}
\label{sec:sft}

We train the policy to imitate repair trajectories weighted by outcome quality, exploration, and agent-step count. The training interface matches the inference interface. Some examples rely on verifier outcomes rather than gold labels, so the corpus can contain incorrect repairs.

\paragraph{Corpus construction.} We curate the SFT corpus from the $230$ issues selected after the all-$500$ failure analysis in Section~\ref{sec:tts}. Trajectories from the remaining $270$ issues are excluded from SFT and RL training. The corpus includes the six search procedures in Table~\ref{tab:efficiency-all} and the methodology-diversity runs, generated with Qwen3-Coder-30B-A3B-Instruct under mini-SWE-agent. Each example records the conversation, final patch, edit locations, agent-step count, verifier outcome, available gold label, and behavior tags. Trajectories are ranked within each \texttt{instance\_id} by the weight defined below, with ties resolved by fewer steps. The retained corpus contains $5{,}061$ examples from $230$ issues.

The corpus includes contrastive-edit examples pairing a correct repair with an incorrect candidate at the same location, and reproduction examples distinguishing buggy from fixed behavior. Contrastive examples receive additional weight when the incorrect candidate passed the verifier. These examples are included in the $5{,}061$ total.

\paragraph{Quality tiers.} Each example receives the weight of the first tier it matches. Tier $T1$ has weight $1.0$ and requires a gold-resolved patch that was selected or received a verifier score of at least $1.0$. Tier $T2$ has weight $0.6$ and requires either a verifier score of at least $1.0$ or a gold-resolved patch with a verifier pass. Tier $T3$ has weight $0.25$ and requires a non-empty patch with a weaker positive verifier signal. All remaining examples receive weight $0$. Only $T1$ requires a gold label.

\paragraph{Exploration and efficiency weights.} We multiply the tier weight by exploration and efficiency factors.
\begin{equation}
  \label{eq:reward-weight}
  w \;=\; \underbrace{w_{\text{tier}}}_{\text{quality}}\cdot
  \underbrace{\bigl(1+\lambda\,\tilde\nu\bigr)}_{\text{exploration}}\cdot
  \underbrace{\Bigl(1+\beta\bigl[1-\min(1,\tfrac{\text{steps}}{S_{\text{ref}}})\bigr]\Bigr)}_{\text{efficiency}},
  \qquad \lambda=1.0,\ \beta=0.5,\ S_{\text{ref}}=120.
  \end{equation}

The effective novelty score $\tilde\nu$ has behavior-dependent floors of $0.6$ for post-ban recovery and self-correction, $0.5$ for localization-win and synthesis-only solves, and $0.4$ for methodology-unique solves. The efficiency factor decreases from $1.5$ at zero steps to $1.0$ at $S_{\text{ref}}=120$ steps and remains $1.0$ for longer trajectories. These factors adjust the contribution of each example to supervised training.

\subsection{Reinforcement Learning for Discriminative Verification}
\label{sec:rl}

The SFT objective imitates trajectories without ever rewarding a test for rejecting an incorrect repair \citep{shum2025swerm}. We add a reinforcement objective that evaluates generated tests against incorrect programs, which measures which incorrect behaviors a test detects.

\paragraph{Generated objects.} For each training issue, the policy produces a patch $P$ and a test $T$ containing assertions $\{a_1,\ldots,a_n\}$. We draw $k$ rollouts per issue. The other $k-1$ patches in the group are sibling repairs, which provide additional candidates for test evaluation. A gold execution oracle $\mathcal G$ determines whether a patch satisfies the training issue's evaluation tests.

\paragraph{Oracle access and rollout feedback.} The policy observes repository files, command results, and the \texttt{pytest} output of $T$ on $P$. Revision rounds return surviving patch variants as counterexamples that the policy can use to revise its patch or test. The policy does not receive gold tests, gold verdicts, assertion-credit scores, or group advantages. Oracle labels are used by the scoring procedure described below. The completed rollout receives a scalar reward, from which we compute trajectory and assertion-token advantages for the training update.

\paragraph{Mutation evaluation.} The harness generates variants from the current patch using deterministic operators and a frozen model distinct from the policy. The operators cover boundary, value, logic, control-flow, and effect faults. Model-generated variants are produced during the run rather than retrieved from a cache. A line trace checks that the test executes the patched lines. A variant counts as detected when an assertion fails. Uncaught errors receive no detection credit.

The scoring set $S$ contains oracle-rejected variants and oracle-rejected sibling repairs. Candidates that pass the gold tests are excluded from this negative set. Passing the gold tests does not establish semantic equivalence, but it prevents those candidates from being scored as known incorrect repairs. We record the detected subset as $\mathrm{Kill}(T)$. Appendix~\ref{app:mutloop} gives a worked example.

The repair loop allows up to three revision rounds and applies a penalty for additional rounds. Exploration terms measure group diversity across edited files and assertion targets, coverage of fault classes, and repetition of test templates. These terms supplement the patch-correctness and test-discrimination signals.

\paragraph{Post-rollout reward.} We score each completed rollout with
\begin{equation}
R_{\text{traj}} \;=\; w_c\,\textsc{Correct} \;+\; w_v\,\textsc{Verif} \;+\; w_e\,\textsc{Edit}
\;+\; w_p\,\textsc{Process},
\label{eq:traj-reward}
\end{equation}

We use $w_c=3.0$, $w_v=1.0$, $w_e=0.4$, and $w_p=0.2$. The correctness term is $1$ when $P$ satisfies $\mathcal G$ and $0$ otherwise. The verification term measures detection of the scored negatives and agreement with the oracle.
\begin{equation}
\textsc{Verif} \;=\; \underbrace{\tfrac{|\mathrm{Kill}(T)|}{|S|}}_{\text{dense slope}}
\;+\; \underbrace{\lambda_{\text{fp}}\,\mathbf{1}[\mathrm{Kill}(T)=S]}_{\text{fixpoint bonus}}
\;+\; \underbrace{h(T)}_{\text{honesty}}
\label{eq:verif}
\end{equation}

The detection fraction provides partial credit, and $\lambda_{\text{fp}}=0.5$ adds a bonus when the test detects every variant in $S$. The agreement term $h(T)$ is $-0.5$ for false acceptance, $-0.25$ for false rejection, and $0$ when the test verdict agrees with the oracle. The resulting verification term lies in $[-0.5,1.5]$. The edit-placement and reproduce-first terms lie in $[0,1]$ and contribute only when $\textsc{Verif}>0$ or $\textsc{Correct}=1$.

The weighted auxiliary terms have a combined range of at most $2.6$, below the correctness weight of $3.0$. Every correct rollout therefore receives a higher scalar reward than every incorrect rollout. The auxiliary terms determine the ordering within each correctness class. This ordering concerns the scalar reward, not the sign of every group-relative advantage.

Detecting all scored variants does not establish correctness beyond that set. For a fixed test and $M$ independent draws from a fixed mutant distribution with no survivors, the one-sided $95\%$ upper confidence bound on the survival probability is $\rho\leq 1-0.05^{1/M}$. Table~\ref{tab:rigor-fixpoint} reports training completion and held-out detection. Appendix~\ref{app:rl-details} specifies empty negative sets, invalid executions, and other boundary cases.

\paragraph{Assertion credit.} We optimize with Group Relative Policy Optimization (GRPO). Rewards normalized within each issue's rollout group give the trajectory advantage $A_{\text{traj}}$. The kill matrix $K_{m,i}$ records whether assertion $i$ detects mutant $m$. We allocate credit to detecting assertions using
\begin{equation}
\tilde c_i \;=\; \sum_{m:\,K_{m,i}=1}\frac{\omega_{\text{cls}(m)}}{\kappa_m},
\qquad c_i \;=\; \frac{\tilde c_i}{\sum_j \tilde c_j}
\label{eq:credit}
\end{equation}

Here, $\kappa_m$ is the number of assertions that detect mutant $m$, and $\omega_{\text{cls}(m)}$ is its fault-class weight. Each detected mutant divides one class-weighted unit among its detecting assertions. Normalization gives $c_i$. We set $s_t=c_{i(t)}$ for tokens in assertion $i(t)$ and $s_t=0$ elsewhere. The token advantage is
\begin{equation}
A_t \;=\; A_{\text{traj}} \;+\; \beta\,G\,A_{\text{traj}}^{+}\,s_t,
\qquad A_{\text{traj}}^{+}=\max(A_{\text{traj}},0),\quad G\in\{0,1\}
\label{eq:shaped-advantage}
\end{equation}

The gate $G=1$ when the patch resolves the issue or the test detects every scored mutant. The assertion bonus is applied when this gate is satisfied and $A_{\text{traj}}>0$. Complete detection can therefore enable assertion credit for an incorrect patch. The bonus rewards detection of scored negatives and does not certify the accompanying patch. The trajectory-level agreement term penalizes false rejection, but the unsigned assertion credit contains no penalty for rejecting correct alternatives. Patch tokens retain the trajectory advantage. Setting $\beta=0$ removes the assertion bonus.

The unnormalized credit $\tilde c_i$ is the Shapley allocation of a weighted coverage game in which an assertion set receives the total weight of the mutants it detects. This interpretation describes the credit allocation and provides no convergence guarantee. Appendix~\ref{app:rl-details} discusses assertion execution and token-span mapping. We use GRPO group size $k=6$, PPO clip $0.2$, and no KL penalty. Appendix~\ref{app:rigor-repro} lists the training settings.

\section{Evaluation}
\label{sec:eval}

\subsection{Evaluation Protocol}
\label{sec:eval-protocol}

\paragraph{Populations and exposure.} Section~\ref{sec:tts} analyzes all $500$ SWE-bench Verified issues. We curate SFT and RL data from $230$ of them and evaluate the trained policies on the remaining $270$, of which $180$ come from training repositories and $90$ from repositories excluded from training. Held-out means excluded from SFT and RL training. All $270$ appeared in the diagnostic, so this split measures held-out issues within a benchmark used for method development rather than an out-of-distribution evaluation. Transfer uses DeepSWE ($113$ tasks), SWE-PolyBench, SWE-bench Lite, and SWE-bench Pro ($731$ issues). The $30$B study evaluates all four after SFT and all but Lite after RL, while the $7$B and $14$B runs also evaluate Lite (Appendix~\ref{app:scale}).

\paragraph{Decoding and aggregation.} The policies share the backbone, scaffold, tools, context limits, and attempt budget. Pass@$1$ uses the greedy first attempt, and eight-attempt evaluation adds seven samples at $T=0.7$, so the samples are not identically distributed. This budget differs from the eight methodology-conditioned chains and the candidate grid of Section~\ref{sec:tts}. The configuration specifies three seeds.

\paragraph{Candidate populations and metrics.} Coverage and selection use each policy's own eight-candidate sets. Verifier discrimination instead uses one frozen bank of labeled repairs per held-out issue, shared across policies, combining natural incorrect repairs from several generators with correct alternatives, with labels withheld from the test-generating policy. Cross-error-source evaluation separates familiar operators, held-out operators, an excluded semantic-mutant generator, and natural incorrect repairs. Repair correctness is defined by the benchmark grader, so resolution and verifier labels measure agreement with its tests rather than semantic correctness. Pass@$8$ is oracle coverage, Selected@$8$ is the fraction whose selected candidate is correct, and selection efficiency is Selected@$8$/pass@$8$. Taking a correct repair as the positive class, precision is $TP/(TP+FP)$, the false-positive rate is $FP/(FP+TN)$, correct acceptance is $TP/(TP+FN)$, and balanced accuracy averages correct acceptance and incorrect rejection. DeepSWE verifier labels use fail-to-pass satisfaction. We report $\eta_k=U_k^{\mathrm{loc}}/k$ as location diversity per attempt, and agent-steps count agent interactions rather than tokens or execution time. Appendix~\ref{app:rigor-defs} defines acceptance coverage, correct-alternative acceptance, and the zero-survivor audit.

\subsection{Repair Accuracy}
\label{sec:eval-sft}
\label{sec:eval-rl}

On the Verified split held out from training, SFT increases pass@$1$ from $31.9\%$ to $35.2\%$ and RL from SFT raises it to $43.0\%$ (Table~\ref{tab:main-rigor}). Applying the same RL objective to the base policy yields $35.9\%$ (Table~\ref{tab:rl-main}), so the objective gains $4.0$ points from the base policy and $7.8$ from the supervised one.

\subsection{Candidate Coverage and Selection}
\label{sec:eval-selection}

SFT increases oracle pass@$8$ from $46.7\%$ to $51.1\%$, while RL raises it to $60.7\%$. Selected@$8$ increases from $33.5\%$ to $38.5\%$ after SFT and to $50.5\%$ after RL. Selection efficiency rises from $0.72$ to $0.75$ and then to $0.83$ (Table~\ref{tab:e6-selector}). The gains therefore involve both the availability of correct repairs and the fraction selected by the verifier. The coverage half traces to the exploratory behavior the corpus weighting teaches. Trained to spread attempts across locations and vary the edit, the policy reaches $1.98$ distinct localizations per issue against $1.41$ for the base model, and the share of samples contributing a new fault location rises from $0.18$ to $0.25$ at eight attempts, so more of the budget lands on ground no earlier sample covered.

\paragraph{Behavioral changes under supervision.} Supervision raises pass@$1$ on every suite that resolves any task, by $3.3$ points on Verified and $3.8$, $8.4$, and $13.4$ on SWE-PolyBench, Lite, and Pro (Table~\ref{tab:all-benchmarks}). The trajectory measures move in the same direction on every suite, with more files read before the first edit, smaller patches, fewer test-file edits, and fewer steps to first submission. Reproduce-first is the one measure where base leads, $94.4\%$ against $92.1\%$ on Verified, while rising on all four transfer suites. This is the behavior the weighting targets. A policy that commits to one site keeps editing it, so a wrong first localization is unrecoverable, and Section~\ref{sec:tts} showed that breadth alone loses to blind retry. The weighting rewards moving on once execution rules a site out, so the added breadth lands where feedback directs it rather than across arbitrary sites. At matched corpus and budget, removing the exploration factor lowers distinct localizations from $1.98$ to $1.74$ and pass@$1$ from $35.2\%$ to $34.8\%$ on Verified, and from $4.13$ to $3.70$ and $24.2\%$ to $23.2\%$ on SWE-PolyBench (Table~\ref{tab:e3-sft-weighting}). Staging the terms raises localization and resolution together, from $1.55$ and $33.7\%$ under uniform weighting to $1.90$ and $34.8\%$ with exploration, while an interface-compliance control reaches $1.45$ and $32.6\%$, below uniform on both (Table~\ref{tab:rigor-sft-ablation}). Which trajectory measure carries the effect is not separated.

Steps to first submission decrease from $62.5$ to $56.0$ (Table~\ref{tab:all-benchmarks}), and $\eta_8$ rises from $0.34$ to $0.61$ on DeepSWE. Location diversity does not measure candidate correctness, so it complements rather than replaces the coverage analysis.

\begin{table}[t]
\caption{Base and SFT results across five evaluation suites. Behavioral measures summarize the evaluated trajectories, diversity measures compare eight attempts per issue, and downward arrows mark metrics where smaller is preferred. Behavioral changes are descriptive rather than independent measures of correctness.}
\label{tab:all-benchmarks}
\centering
\renewcommand{\arraystretch}{0.70}
\resizebox{\textwidth}{!}{%
\begin{tabular}{l Y O Y O Y O Y O Y O}
\toprule
\hdr
& \multicolumn{2}{c}{SWE-bench Verified} & \multicolumn{2}{c}{DeepSWE}
& \multicolumn{2}{c}{SWE-PolyBench} & \multicolumn{2}{c}{SWE-bench Lite}
& \multicolumn{2}{c}{SWE-bench Pro} \\
\cmidrule(lr){2-3}\cmidrule(lr){4-5}\cmidrule(lr){6-7}\cmidrule(lr){8-9}\cmidrule(lr){10-11}
\hdr
Metric & Base & SFT & Base & SFT & Base & SFT & Base & SFT & Base & SFT \\
\midrule
\grp \multicolumn{11}{l}{\textit{Resolution rate}} \\
\midrule
pass@1                                & $31.9$ & \wc{$\mathbf{35.2}$} & $0.0$ & $0.0$ & $20.4$ & \wc{$\mathbf{24.2}$} & $28.3$ & \wc{$\mathbf{36.7}$} & $2.3$ & \wc{$\mathbf{15.7}$} \\
pass@8                                & $46.7$ & \wc{$\mathbf{51.1}$} & $0.0$ & $0.0$ & $24.6$ & \wc{$\mathbf{30.2}$} & $30.7$ & \wc{$\mathbf{39.7}$} & $7.4$ & \wc{$\mathbf{19.6}$} \\
\midrule
\grp \multicolumn{11}{l}{\textit{Exploration}} \\
\midrule
Files read before first edit          & $1.87$ & \wc{$\mathbf{2.19}$} & $1.26$ & \wc{$\mathbf{2.12}$} & $2.69$ & \wc{$\mathbf{3.91}$} & $2.61$ & \wc{$\mathbf{3.89}$} & $4.71$ & \wc{$\mathbf{6.81}$} \\
Read-to-edit ratio                    & $3.44$ & \wc{$\mathbf{7.49}$} & $9.14$ & \wc{$\mathbf{13.7}$} & $6.21$ & \wc{$\mathbf{8.19}$} & $7.24$ & \wc{$\mathbf{9.50}$} & $5.56$ & \wc{$\mathbf{7.61}$} \\
Search commands / trajectory          & $28.9$ & \wc{$\mathbf{33.5}$} & $48.3$ & \wc{$\mathbf{52.8}$} & $25.9$ & \wc{$\mathbf{30.1}$} & $28.2$ & \wc{$\mathbf{33.8}$} & $14.5$ & \wc{$\mathbf{18.8}$} \\
\midrule
\grp \multicolumn{11}{l}{\textit{Reproduction \& testing}} \\
\midrule
Reproduce-first rate                  & $\mathbf{94.4}$ & $92.1$ & $28.7$ & \wc{$\mathbf{31.9}$} & $23.3$ & \wc{$\mathbf{27.1}$} & $15.7$ & \wc{$\mathbf{16.9}$} & $7.1$ & \wc{$\mathbf{12.5}$} \\
Test-execution rate                   & $92.1$ & $92.1$ & $24.4$ & \wc{$\mathbf{27.8}$} & $86.2$ & \wc{$\mathbf{88.2}$} & $96.3$ & \wc{$\mathbf{98.2}$} & $31.1$ & \wc{$\mathbf{37.8}$} \\
Test-after-fix rate                   & $91.9$ & \wc{$\mathbf{93.1}$} & $23.5$ & \wc{$\mathbf{25.1}$} & $84.5$ & \wc{$\mathbf{87.6}$} & $91.3$ & \wc{$\mathbf{93.3}$} & $11.1$ & \wc{$\mathbf{18.5}$} \\
\midrule
\grp \multicolumn{11}{l}{\textit{Self-correction}} \\
\midrule
Self-correction rate                  & $77.9$ & \wc{$\mathbf{81.2}$} & $17.3$ & \wc{$\mathbf{21.8}$} & $44.8$ & \wc{$\mathbf{45.7}$} & $47.0$ & \wc{$\mathbf{53.1}$} & $13.7$ & \wc{$\mathbf{20.1}$} \\
\midrule
\grp \multicolumn{11}{l}{\textit{Edit discipline}} \\
\midrule
Patch size, lines changed $\downarrow$ & $4$ & \wc{$\mathbf{2}$} & $4$ & \wc{$\mathbf{2}$} & $4$ & \wc{$\mathbf{2}$} & $4$ & \wc{$\mathbf{2}$} & $71$ & \wc{$\mathbf{60}$} \\
Source files touched                  & $1$ & $1$ & $1$ & $1$ & $1$ & $1$ & $1$ & $1$ & $1$ & $1$ \\
Test-file-edit rate $\downarrow$       & $30.2$ & \wc{$\mathbf{18.4}$} & $13.7$ & \wc{$\mathbf{8.80}$} & $11.2$ & \wc{$\mathbf{7.4}$} & $17.3$ & \wc{$\mathbf{12.1}$} & $9.3$ & \wc{$\mathbf{5.5}$} \\
\midrule
\grp \multicolumn{11}{l}{\textit{Efficiency}} \\
\midrule
Steps to first submission $\downarrow$ & $62.5$ & \wc{$\mathbf{56.0}$} & $84.9$ & \wc{$\mathbf{76.5}$} & $54.2$ & \wc{$\mathbf{46.5}$} & $61.5$ & \wc{$\mathbf{55.5}$} & $29.5$ & \wc{$\mathbf{21.5}$} \\
Format/context-overflow rate $\downarrow$ & $20.3$ & \wc{$\mathbf{10.4}$} & $30.8$ & \wc{$\mathbf{18.6}$} & $27.6$ & \wc{$\mathbf{16.3}$} & $22.7$ & \wc{$\mathbf{10.8}$} & $39.4$ & \wc{$\mathbf{25.7}$} \\
\midrule
\grp \multicolumn{11}{l}{\textit{Solution diversity (across 8 samples)}} \\
\midrule
Distinct localizations                & $1.41$ & \wc{$\mathbf{1.98}$} & $2.74$ & \wc{$\mathbf{4.86}$} & $3.33$ & \wc{$\mathbf{4.13}$} & $2.72$ & \wc{$\mathbf{3.81}$} & $3.77$ & \wc{$\mathbf{4.51}$} \\
Distinct patch signatures             & $5.77$ & \wc{$\mathbf{6.10}$} & $4.26$ & \wc{$\mathbf{5.11}$} & $3.12$ & \wc{$\mathbf{3.89}$} & $3.21$ & \wc{$\mathbf{5.62}$} & $6.37$ & \wc{$\mathbf{7.54}$} \\
\bottomrule
\end{tabular}%
}
\par\smallskip
{\footnotesize Format/context-overflow rate is the fraction of rollouts with at least one recoverable protocol or context event. An event can occur in a rollout that resolves the issue.}

\end{table}

\begin{table}[t]
\centering
\small
\setlength{\tabcolsep}{5pt}
\begin{tabular}{l YY Y Y Y}
\toprule
\hdr
\textbf{Policy} & \textbf{Oracle@8} & \textbf{Selected@8} & \textbf{Majority vote} & \textbf{Random pick} & \textbf{Sel.\ eff.} \\
\midrule
Base                 & $46.7$ & $33.5$ & $30.0$ & $26.0$ & $0.72$ \\
SFT                  & $51.1$ & $38.5$ & $35.0$ & $30.2$ & $0.75$ \\
\winrow
\textbf{$+$RL}       & $60.7$ & $50.5$ & $44.0$ & $36.0$ & $0.83$ \\
\bottomrule
\end{tabular}
\caption{Coverage and selection on the Verified split held out from training using the existing eight-candidate sets. Oracle@$8$ is the fraction of issues with a correct candidate. Selected@$8$ is the fraction for which the verifier returns a correct candidate. Selection efficiency is their ratio. No additional repairs are generated for this analysis.}
\label{tab:e6-selector}
\end{table}

\subsection{Verifier Discrimination on Fixed Candidates}
\label{sec:eval-discrimination}

\begin{table}[t]
\centering
\small
\setlength{\tabcolsep}{5pt}
\renewcommand{\arraystretch}{0.85}
\resizebox{\textwidth}{!}{%
\begin{tabular}{l YY Y Y Y Y}
\toprule
\hdr
\textbf{Verifier policy} & \textbf{Wrong rej.\ (TNR)} & \textbf{Correct acc.\ (TPR)} & \textbf{Bal.\ acc.} & \textbf{Accept.\ cov.} & \textbf{Sel.\ acc.} & \textbf{FPR $\downarrow$} \\
\midrule
Base                         & $70.0$ & $68.0$ & $69.0$ & $66.0$ & $52.0$ & $30.0$ \\
SFT                          & $73.0$ & $72.0$ & $72.5$ & $70.0$ & $58.0$ & $27.0$ \\
Correctness-only RL          & $78.0$ & $74.0$ & $76.0$ & $73.0$ & $63.0$ & $22.0$ \\
\winrow
\textbf{Full ($+$RL)}        & $87.4$ & $80.0$ & $83.7$ & $78.0$ & $72.0$ & $12.6$ \\
\bottomrule
\end{tabular}%
}
\caption{Verifier discrimination on a frozen bank of labeled repairs shared across policies. The bank contains natural incorrect repairs from multiple generators and correct alternatives. Labels are not exposed to the test-generating policy. Wrong rejection and correct acceptance measure the two sides of discrimination. Metric denominators are defined in Section~\ref{sec:eval-protocol} and Appendix~\ref{app:rigor-defs}.}
\label{tab:rigor-fixed-bank}
\end{table}

\begin{wrapfigure}{r}{0.4\linewidth}
\centering
\includegraphics[width=\linewidth]{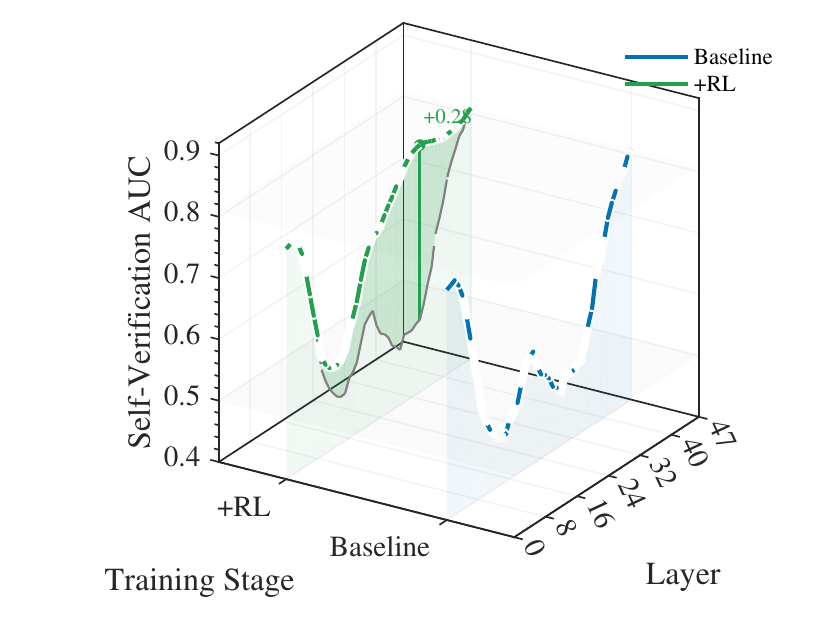}
\caption{Layer-wise self-verification AUC across training stages. The post-RL representation has higher AUC for distinguishing correct and incorrect patches, with the largest differences in the mid-to-late layers (shaded).}
\label{fig:self_verification_3d}
\end{wrapfigure}

On one frozen candidate bank shared across policies, so that differences reflect the verifier and not an easier candidate population, the RL verifier rejects $87.4\%$ of incorrect repairs against $73.0\%$ for SFT, while correct-patch acceptance and balanced accuracy rise with it, so both sides of the decision improve rather than the verifier becoming uniformly stricter. Correctness-only RL captures roughly a third of that gain (Table~\ref{tab:rigor-fixed-bank}). The per-assertion credit pushes the test toward assertions that each catch something the others miss, since a detected variant divides one unit among its detectors and a redundant assertion collects less than a new one. Two mutant sources keep that pressure from collapsing onto one distribution, since an assertion set tuned to operator shapes leaves model-generated variants alive. Training on operators alone loses pass@$1$, precision, and zero-survivor rate together (Table~\ref{tab:rl-adversary}), and rejection degrades gradually off the training distribution (Table~\ref{tab:rigor-transfer}).

The verification code itself changes. The median test carries no assertions before training and $11$ after, length falls from $42$ to $33$ lines, and it covers $3$ of $5$ fault classes rather than none (Table~\ref{tab:rl-main}). A test with no assertions cannot reject any variant.

Holding the base repair generator fixed, replacing its verifier with the trained verifier raises resolution from $31.9\%$ to $36.5\%$ (Table~\ref{tab:rigor-swap}). The estimated interaction between changing the generator and changing the verifier is $-0.1$ percentage points, with a $95\%$ confidence interval of $[-2.6,2.4]$. This experiment detects a verifier contribution without establishing a nonzero interaction.

Figure~\ref{fig:self_verification_3d} reports a layer-wise probe comparison. The post-RL curve has higher AUC, with its largest differences in the mid-to-late layers. This association does not establish a causal contribution to repair accuracy.

\subsection{Reward Controls}
\label{sec:eval-controls}

The RL controls start from one SFT checkpoint and match training issues, update counts, and rollout budgets. Correctness-only GRPO reaches $39.6\%$ pass@$1$ and $28.5\%$ verifier precision. Adding scalar verification reward raises these values to $40.7\%$ and $38.4\%$. Assertion credit raises them to $43.0\%$ and $41.7\%$ (Table~\ref{tab:main-rigor}b). Both the verification reward and assertion credit contribute to the reported gain.

Uniform assertion credit yields $38.0\%$ precision and shuffling it at matched total mass yields $31.5\%$, so placement affects discrimination beyond the amount assigned. Removing mutation evaluation reduces pass@$1$ from $43.0\%$ to $38.1\%$, the largest reduction among the component removals in Table~\ref{tab:rl-ablations}. Allowing uncaught errors to count as detections reduces it to $39.6\%$. These removals measure contributions within the full configuration and need not have additive effects. Ungating the shaping terms costs $1.9$ points, removing the repair loop costs $3.0$ while leaving precision at $40.1\%$, and removing the exploration terms costs $1.5$ (Table~\ref{tab:rl-ablations}).

Group size raises both metrics, from $41.5\%$ pass@$1$ and $39.0\%$ precision at $k{=}4$ to $43.0\%$ and $41.7\%$ at the trained $k{=}6$ and $43.3\%$ and $42.6\%$ at $k{=}16$. Lowering the correctness weight raises precision to $43.2\%$ while pass@$1$ falls to $38.9\%$ (Table~\ref{tab:rl-sensitivity}), so the two respond differently to the same reward change.

\begin{table}[t]
\centering
\caption{Training exposure and reward controls. Panel (a) separates the $230$ training issues, $180$ test issues from training repositories, $90$ test issues from repositories excluded from training, and SWE-bench Pro. Panel (b) compares objectives from one SFT checkpoint at matched training issues and update and rollout budgets. $\Delta$ is the pass@$1$ difference from the full objective in percentage points.}
\label{tab:main-rigor}
{\footnotesize\textbf{(a) Resolution by training exposure}}\\[2pt]
\footnotesize
\renewcommand{\arraystretch}{0.92}
\setlength{\tabcolsep}{4pt}
\begin{tabular}{l c YOO YOO}
\toprule
\hdr
& & \multicolumn{3}{c}{\textbf{pass@1}} & \multicolumn{3}{c}{\textbf{pass@8}} \\
\cmidrule(lr){3-5}\cmidrule(lr){6-8}
\hdr
\textbf{Evaluation subset} & \textbf{$N$} & \textbf{Base} & \textbf{SFT} & \textbf{+RL} & \textbf{Base} & \textbf{SFT} & \textbf{+RL} \\
\midrule
Train set, Verified (reference)       & $230$ & $33.9$ & $37.4$ & $45.7$ & $49.1$ & $53.5$ & $63.5$ \\
Test issues, training repositories & $180$ & $33.3$ & $36.7$ & $44.4$ & $48.9$ & $53.3$ & $62.8$ \\
Test issues, nontraining repositories             & $90$  & $28.9$ & $32.2$ & $40.0$ & $42.2$ & $46.7$ & $56.7$ \\
Frozen external (SWE-bench Pro)       & $731$ & $2.3$  & $15.7$ & $20.2$ & $7.4$  & $19.6$ & $22.8$ \\
\midrule
\grp \multicolumn{2}{l}{All $270$ held-out test (Verified)} & $31.9$ & $35.2$ & $43.0$ & $46.7$ & $51.1$ & $60.7$ \\
\bottomrule
\end{tabular}

\vspace{3pt}
{\footnotesize\textbf{(b) Reinforcement-learning controls (from one shared SFT checkpoint)}}\\[2pt]
\setlength{\tabcolsep}{5pt}
\begin{tabular}{l YY Y Y c}
\toprule
\hdr
\textbf{Configuration} & \textbf{pass@1} & \textbf{pass@8} & \textbf{Verif.\ prec.} & \textbf{Correct-alt acc.} & \textbf{$\Delta$ vs full} \\
\midrule
SFT initialization (no RL)                       & $35.2$ & $51.1$ & $26.8$ & $60.0$ & $-7.8$ \\
Correctness-only GRPO                             & $39.6$ & $56.7$ & $28.5$ & $64.0$ & $-3.4$ \\
Correctness $+$ scalar \textsc{Verif} ($\beta{=}0$) & $40.7$ & $58.5$ & $38.4$ & $72.0$ & $-2.3$ \\
\winrow
\textbf{Full objective ($+$ token credit)}        & $43.0$ & $60.7$ & $41.7$ & $78.0$ & $-$ \\
Matched uniform assertion credit                  & $41.5$ & $59.6$ & $38.0$ & $73.0$ & $-1.5$ \\
Shuffled-credit control (matched mass)            & $40.0$ & $57.8$ & $31.5$ & $64.0$ & $-3.0$ \\
Prior mutation-RL control               & $40.7$ & $58.1$ & $35.5$ & $60.0$ & $-2.3$ \\
\bottomrule
\end{tabular}
\end{table}

\subsection{Transfer and Limitations}
\label{sec:eval-transfer}

RL improves pass@$1$ over SFT by $8.3$ percentage points on train-exposed issues, $7.7$ on test issues from training repositories, and $7.8$ on repositories excluded from training, reaching $45.7\%$, $44.4\%$, and $40.0\%$ respectively (Table~\ref{tab:main-rigor}a). The gain is near-constant across the exposure gradient while the absolute level falls, the pattern expected if the objective transfers rather than recalling training fixes. On SWE-PolyBench, pass@$1$ increases from $24.2\%$ to $31.8\%$. On SWE-bench Pro, it increases from $15.7\%$ to $20.2\%$ (Table~\ref{tab:rl-benchmarks}). These evaluations support transfer beyond the training issues. They do not exclude pretraining exposure to public repositories.

DeepSWE full resolution remains zero at this model scale, but its internal measures improve. SFT raises the tasks with at least one satisfied fail-to-pass test from $17$ to $21$ of $113$ and the mean best-of-eight satisfied fraction from $4.0\%$ to $4.8\%$ (Table~\ref{tab:deepswe}), which is partial test satisfaction rather than completed repair. Verifier precision uses a different positive criterion here (Section~\ref{sec:eval-protocol}).

\begin{table}[t]
\centering
\small
\renewcommand{\arraystretch}{0.9}
\resizebox{\textwidth}{!}{%
\begin{tabular}{l YOYO YOYO YOYO}
\toprule
\hdr
& \multicolumn{4}{c}{DeepSWE}
& \multicolumn{4}{c}{SWE-PolyBench} & \multicolumn{4}{c}{SWE-bench Pro} \\
\cmidrule(lr){2-5}\cmidrule(lr){6-9}\cmidrule(lr){10-13}
\hdr
Metric & Base & \textbf{+RL} & SFT & \textbf{+RL}
       & Base & \textbf{+RL} & SFT & \textbf{+RL}
       & Base & \textbf{+RL} & SFT & \textbf{+RL} \\
\midrule
pass@1                    & $0.0$ & $0.0$ & $0.0$ & $0.0$ & $20.4$ & \wc{$23.8$} & $24.2$ & \wc{$31.8$} & $2.3$  & \wc{$15.5$} & $15.7$ & \wc{$20.2$} \\
pass@8                    & $0.0$ & $0.0$ & $0.0$ & $0.0$& $24.6$ & \wc{$30.2$} & $30.2$ & \wc{$41.6$} & $7.4$  & \wc{$19.4$} & $19.6$ & \wc{$22.8$} \\
Verifier precision        & $21.1$   & \wc{$33.7$} & $23.1$& \wc{$38.6$} & $13.6$    & \wc{$22.8$} & $17.5$ & \wc{$26.8$}& $15.5$    & \wc{$24.5$} & $21.0$ & \wc{$24.7$} \\
False-positive rate $\downarrow$    & $23.0$& \wc{$14.6$} & $21.1$& \wc{$11.1$} & $19.0$ & \wc{$14.6$} & $18.1$ & \wc{$11.3$} & $21.7$ & \wc{$14.5$}& $20.9$ & \wc{$12.7$}\\
Fixpoint (zero-survivor) rate & $3.0$& \wc{$3.9$} & $5.11$& \wc{$8.3$} & $0.4$ & \wc{$2.1$} & $3.87$ & \wc{$7.31$} & $1.90$ & \wc{$4.51$} & $2.81$ & \wc{$4.19$} \\
\bottomrule
\end{tabular}%
}
\caption{Transfer evaluation from base and SFT initializations. Each initialization is paired with its RL-trained policy. DeepSWE verifier precision uses a fail-to-pass criterion rather than full task resolution.}
\label{tab:rl-benchmarks}
\end{table}

The signed assertion-credit ablation raises correct-alternative acceptance from $78.0\%$ to $84.0\%$, cross-generator discrimination from $74.0\%$ to $78.0\%$, and pass@$1$ from $43.0\%$ to $43.3\%$ (Table~\ref{tab:rigor-signed-credit}). This result supports adding an assertion-level penalty for false rejection. The main results use unsigned credit, so the signed variant's gains apply to this ablation.

Verification remains sensitive to the source of incorrect repairs. The RL verifier rejects $91.0\%$ of familiar-operator mutants, $84.0\%$ of held-out operators, $79.0\%$ of variants from an excluded generator, and $74.0\%$ of natural incorrect repairs from independent generators. The latter exceeds SFT by $12$ percentage points, while correct-alternative acceptance remains $78.0\%$ (Table~\ref{tab:rigor-transfer}). Complete detection of a finite mutant set does not establish correctness outside that set.

The $7$B and $14$B experiments gain from both stages. Verified pass@$8$ rises $21.9\to30.4\to33.3\%$ at $7$B and $27.0\to36.7\to42.6\%$ at $14$B, and SWE-PolyBench rises $8.0\to17.8\%$ and $19.8\to37.4\%$ (Table~\ref{tab:scale-trend}). Published coder results in Appendix~\ref{app:coder-baselines} provide context but use different scaffolds, subsets, or selection protocols. They do not constitute matched comparisons. Agent-step savings also exclude differences in token lengths and test-execution costs.

\section{Related Work}
\paragraph{Repository search and feedback.} SWE-agent and OpenHands provide interfaces for repository inspection, editing, and execution \citep{yang2024sweagent,wang2024openhands}. Agentless organizes repair around localization \citep{xia2024agentless}, and inference-time scaling studies examine additional attempts \citep{brown2024monkeys,snell2024scaling}. Our diagnostic holds backbone and scaffold fixed and compares how location search and verification affect repair. Generated tests also support candidate selection \citep{chen2023codet}, and we measure candidate coverage and selected accuracy as distinct outcomes.

\paragraph{Training repair and test generation.} CodeRL trains code generation from execution feedback \citep{le2022coderl}, SWE-Gym studies training repository agents and verifiers \citep{pan2024swegym}, and CURE co-trains a coder and unit tester with reinforcement learning \citep{wang2025cure}. Our policy generates both the patch and its test, and the reward combines patch correctness with detection of oracle-rejected variants. SWT-Bench evaluates tests for repository bug fixes \citep{mundler2024swtbench}, a related setting for assessing tests beyond whether they execute.

\paragraph{Mutation rewards and credit assignment.} MIST-RL rewards each test for detecting mutants earlier tests missed and penalizes redundancy \citep{zhu2026mistrl}. Ockhamareto is the closest comparator for our token-level credit, mapping mutation contributions to test-block tokens under a Pareto gate on suite effectiveness and size \citep{huang2026ockhamareto}. Our credit divides each detected mutant's weight among its detecting assertions and applies a positive-advantage bonus gated by repair correctness or complete detection, with the trajectory reward prioritizing repair correctness. These differences concern the reward and credit rule rather than introducing mutation testing or token-level attribution.

\paragraph{Test and patch revision.} CoHarden generates a reproduction before editing and uses mutant-based feedback to revise tests and fixes \citep{tan2026coharden}, the closest comparison to our revision loop. We also train the policy from those signals and evaluate the resulting verifier on frozen candidates, correct alternatives, and incorrect repairs from independent generators.

\section{Conclusion}
We hypothesized that reliable repair rests on three learnable behaviors, locating the fault, varying the edit, and verifying which candidate to return. The first two determine candidate coverage and the third selection quality, neither compensating for the other, and the swap estimates their contributions as additive. Weighted supervision and mutation-based reinforcement learning improve both, with gains on SWE-bench Pro and SWE-PolyBench and, on DeepSWE, partial credit without full resolution. The constraint is general, since a proposer scoring its own work cannot be wrong, so negatives must be manufactured rather than collected.

Where the verifier is weakest indicates what to improve. Rejection falls from $91.0\%$ on familiar operators to $74.0\%$ on repairs from independent generators, so a wider negative distribution during training is the first direction. The second is the signed credit rule, which charges an assertion for rejecting a correct alternative and raises correct-alternative acceptance from $78.0\%$ to $84.0\%$ in our ablation, though not yet under the full evaluation. Folding the verifier into the search loop, so that it steers exploration rather than filtering afterwards, would reach the coverage limit selection cannot.

\section*{AI Use Statement}
Generative AI was used only to improve the grammar, clarity, and concision of author-written text, and coding assistants were used for routine tasks such as plotting, table generation, and code refactoring. AI was not used to develop the research ideas, design the methodology or experiments, or generate or interpret experimental results. The authors reviewed and verified all AI-assisted content and take full responsibility for the paper.

\section*{Ethics Statement}
This work uses publicly available models and benchmarks and involves no human subjects, personally identifiable information, or new data collection.

\section*{Reproducibility Statement}
Section~\ref{sec:methodology} defines the method, and Section~\ref{sec:eval-protocol} specifies the evaluation protocol. Appendices~\ref{app:rl-details} and~\ref{app:rigor-repro} report implementation details and training hyperparameters. Search rules, prompts, and experiment configurations appear in Appendices~\ref{app:search-details}, \ref{app:tts:methodologies}, and~\ref{app:tts:experiments}. Appendices~\ref{app:rl-ablations} and~\ref{app:rigor} report ablations.

\bibliography{main}
\bibliographystyle{iclr2027_conference}

\appendix
\section{Appendix}
This appendix provides reward ablations, implementation details, comparator studies, and a worked reward example. It also reports transfer and model-scale results and documents the prompts, search logic, and experiment configurations.

\subsection{Test-time search procedures}
\label{app:search-details}

\emph{Self-verification.} The agent submits a patch, then runs up to three audit passes that read the repository, run tests, and revise, with no grader feedback (prompt in Appendix~\ref{app:tts:selfverify}). It resolves $184/500$ ($36.8\%$) at $189{,}371$ agent-steps ($1{,}029$ per solve), yet $78$ to $85\%$ of audit passes change no code and across four attempts it reaches only $1.10$ distinct locations and $1.31$ distinct edits. It neither escapes its wrong first guess nor stops paying to re-examine it, taking both the lowest accuracy and the highest cost.

\emph{Single-axis diagnosis.} One attempt per issue, no retry and no diversity mechanism, fixing the single-sample reference the other procedures answer to. It resolves $189/500$ ($37.8\%$) at $41{,}332$ agent-steps, the cheapest run in the study at $219$ steps per solve.

\emph{Methodology diversity.} Running each issue under eight problem-solving preambles (Appendix~\ref{app:tts:methodologies}) solves overlapping but distinct subsets, with combined oracle coverage $305/500$ ($61.0\%$), above any single preamble. This label-using union is not a deployable rule, yet it marks the headroom. The agent already produces a correct fix for far more issues than one strategy surfaces, so selection limits performance more than generation.

\emph{Directed exploration.} Up-front localization pins one attempt to each of $2.46$ ranked sites yet resolves only $210/500$ ($42.0\%$), below blind retries ($213/500$), so broad coverage alone does not help. Spending attempts on the right sites does, and that needs execution feedback. Banning a confirmed-wrong edit signature and stopping on a strong positive resolves $217/500$ ($43.4\%$) at $131{,}417$ steps, reaching $1.84$ locations and $2.79$ edits with early stops on $65$ issues.

\emph{Reactive location diversity.} Banning the whole location, its files and functions, resolves $225/500$ ($45.0\%$) at $585$ steps per solve, under the ban and early-stop rules of Appendix~\ref{app:tts:ban}. Location diversity pays only when earned reactively, and reactive location-banning ($225$) exceeds reactive edit-banning ($217$), blind retries ($213$), and proactive site-planning ($210$), in that order.

\emph{Differential-verification search.} The combined procedure builds up to eight candidate cells, each pairing one localization with one repair approach. It scores every candidate by executing the agent's own reproduction twice, once on the tree carrying the patch and once on that same tree with the patch reverted. A failure on the reverted tree together with a pass on the patched tree shows the patch changed the behavior the reproduction measures, which is weaker than showing the patch correct elsewhere. The score itself rises with how much independent evidence stands behind that verdict, and it decides whether search continues. A confirmed differential scores $1.2$ when two further checks agree, the repository's existing tests still pass and the reproduction touches the changed lines, and $1.0$ when either is missing, which reads a regression failure as pre-existing rather than caused by the patch. A patch whose reproduction passes while the reverted run returns no verdict scores $0.3$, since the pass is real but nothing corroborates it. No usable signal or a broken harness scores $0$. A vacuous differential, where the reproduction passes on both trees, means it never exercised the fault, so the candidate takes a hard $0$ and the location is pruned. Any score of at least $1.0$ is strong enough to stop the search. Every decision uses the agent's own reproduction and the repository tests, never the hidden grader.

Differential-verification search resolves $264/500$ ($52.8\%$) against $234/500$ ($46.8\%$) for the eight-sample control, at $132{,}697$ agent-steps, $48.1\%$ of the control's $275{,}943$, and $3.12$ attempts per issue. That is $503$ steps per solve against $1{,}179$. Early stopping fires on about $102$ issues and pruning drops $126$ localizations across $73$ issues, so the gains come from spending compute where feedback says it will pay. A component ablation (Appendix~\ref{tab:e9e-dvs}) isolates each part. Dropping the two-tree differential does the most damage ($52.8\to47.4\%$, near the eight-sample baseline), while early stopping costs compute alone (steps/solve $503\to728$ at unchanged resolution).

\subsection{Reinforcement learning results by starting point}
\label{app:rl-main-table}
The reinforcement objective is applied twice, once to the base policy and once to the supervised policy, so that the contribution of the initialization can be read off directly. Both starting points improve on every metric in Table~\ref{tab:rl-main}, which shows the objective does not depend on supervision to function. The size of the gain does depend on it. Pass@$1$ rises by $4.0$ points from the base policy and by $7.8$ from the supervised one, and verifier precision by $9.6$ against $14.9$ points, so the same reward is worth roughly twice as much once the policy already searches widely. The reason is visible in how GRPO forms its advantage, since rollouts that repeat the same edit share the same outcome and leave the group-relative advantage flat, while a wider candidate set puts correct and incorrect patches in the same group and gives the verifier term something to separate. Table~\ref{tab:rigor-fixed-bank} scores the resulting verifiers on one shared bank of labeled repairs, which removes the possibility that the improvement reflects an easier candidate population rather than a better verifier.

\begin{table}[t]
\centering
\small
\renewcommand{\arraystretch}{0.9}
\begin{tabular}{l Y O Y O}
\toprule
\hdr
& \multicolumn{2}{c}{from Base} & \multicolumn{2}{c}{from SFT} \\
\cmidrule(lr){2-3}\cmidrule(lr){4-5}
\hdr
Metric & pre-RL & \textbf{+RL} & pre-RL & \textbf{+RL} \\
\midrule
\grp \multicolumn{5}{l}{\textit{Resolution rate (SWE-bench Verified)}} \\
\midrule
pass@1                         & $31.9$ & \wc{$35.9$} & $35.2$ & \wc{$43.0$} \\
pass@8                         & $46.7$ & \wc{$53.7$} & $51.1$ & \wc{$60.7$} \\
\midrule
\grp \multicolumn{5}{l}{\textit{Verifier quality (vs.\ gold, held-out)}} \\
\midrule
Verifier precision             & $23.5$ & \wc{$33.1$} & $26.8$ & \wc{$41.7$} \\
False-positive rate $\downarrow$ & $30.0$ & \wc{$19.0$} & $27.0$ & \wc{$12.6$}\\
Fixpoint (zero-survivor) rate  & $1.5$ & \wc{$6.3$} & $2.8$ & \wc{$15.6$} \\
\midrule
\grp \multicolumn{5}{l}{\textit{Verification-code discipline}} \\
\midrule
Assertions / test (median)     & 0 & \wc{5} & 0 & \wc{11} \\
Test length, lines $\downarrow$ & 42 & \wc{39} & 42 & \wc{33} \\
Fault classes covered / test   & 0/5 & \wc{3/5} & 0/5 & \wc{3/5} \\
\bottomrule
\end{tabular}
\caption{Base and SFT policies before and after RL on the $270$ Verified issues held out from training. Pass@$1$ measures greedy repair accuracy and pass@$8$ measures oracle coverage. Verifier metrics use oracle correctness labels. Assertion counts, test length, and fault-class coverage describe the generated tests.}
\label{tab:rl-main}
\end{table}

\subsection{Reinforcement Learning Ablations}
\label{app:rl-ablations}

Each ablation removes or varies one component of the objective or training
configuration described in Section~\ref{sec:rl}. Unless otherwise noted,
runs start from the supervised policy and are evaluated on SWE-bench
Verified. We report the issue resolution rate (pass@$1$) and verifier
precision against gold correctness labels. Table~\ref{tab:rl-adversary} varies the adversary that
generates the semantic mutants, and Table~\ref{tab:rl-sensitivity} sweeps the reward weights and the
group size.

\begin{table}[t]
\centering
\small
\begin{tabular}{l YY Y}
\toprule
\hdr
Configuration & pass@1 & Verifier prec. & $\Delta$ vs.\ full \\
\midrule
\winrow
\textbf{Full objective}                            & $43.0$ & $41.7$ & $-$ \\
\quad $\beta=0$ (flat scalar, no step-credit)      & $40.7$ & $38.4$ & $-2.3$ \\
\quad no mutation gauntlet (self-consistency only) & $38.1$ & $29.5$ & $-4.9$ \\
\quad credit crash-kills (no assertion gate)       & $39.6$ & $33.0$ & $-3.4$ \\
\quad ungated $E$, $P$ shaping terms               & $41.1$ & $37.9$ & $-1.9$ \\
\quad no in-episode repair loop                    & $40.0$ & $40.1$ & $-3.0$ \\
\quad no exploration terms                         & $41.5$ & $40.9$ & $-1.5$ \\
\bottomrule
\end{tabular}
\caption{Ablations of the reinforcement objective on SWE-bench Verified, each removing one
component the methodology argues is load-bearing. $\Delta$ is the change in resolution
against the full objective.}
\label{tab:rl-ablations}
\end{table}

\begin{table}[t]
\centering
\small
\begin{tabular}{l YYY}
\toprule
\hdr
\textbf{Semantic-mutant adversary} & \textbf{pass@1} & \textbf{Verifier prec.} & \textbf{Fixpoint rate} \\
\midrule
Weak (same-size, untuned)      & $41.1$ & $37.5$ & $11.8$ \\
Same-family (policy's sibling) & $41.9$ & $39.8$ & $13.9$ \\
\winrow
\textbf{Strong (frozen, ours)} & \wc{$43.0$} & \wc{$41.7$} & \wc{$15.6$} \\
Deterministic mutants only     & $40.4$ & $35.0$ & $\phantom{0}9.5$ \\
\bottomrule
\end{tabular}
\caption{Robustness to the adversary used to generate semantic mutants.
All values are percentages. Every semantic-mutant configuration beats the
deterministic-only baseline on all three metrics, so the benefit extends beyond
the strong adversary, which leads by $2.6$, $6.7$, and $6.1$ points over
deterministic mutants alone.}
\label{tab:rl-adversary}
\end{table}

\begin{table}[t]
\centering
\small
\begin{tabular}{l OY c Y O Y}
\toprule
\hdr
& \multicolumn{2}{c}{\textbf{Reward weights $w_c\!:\!w_v$}} & &
\multicolumn{3}{c}{\textbf{GRPO group size $k$}} \\
\cmidrule(lr){2-3}\cmidrule(lr){5-7}
\hdr
\textbf{Metric} & \textbf{High $w_c$} & \textbf{Low $w_c$} & & \textbf{$k{=}4$} & \textbf{$k{=}6$} & \textbf{$k{=}16$} \\
\midrule
pass@1         & $43.0$ & $38.9$ & & $41.5$ & $43.0$ & \wc{$43.3$} \\
Verifier prec. & $41.7$ & \wc{$43.2$} & & $39.0$ & $41.7$ & \wc{$42.6$} \\
\bottomrule
\end{tabular}
\caption{Sensitivity to the relative correctness and verifier reward
weights ($w_c\!:\!w_v$) and the GRPO group size $k$, the number of rollouts
drawn per training instance. All values are percentages. The trained setting of
Section~\ref{sec:rl} is $w_c{:}w_v = 3.0{:}1.0$ at $k{=}6$, tinted in both
blocks because it is the shared reference point. Lower correctness weighting increases verifier precision by $1.5$
percentage points but reduces pass@1 by $4.1$. Group size pays off early. Moving
from $k{=}4$ to $k{=}6$ adds $1.5$ pass@1 and $2.7$ precision, while going on to
$k{=}16$ adds a further $0.3$ and $0.9$ for more than twice the rollouts per
instance, which is why we train at $k{=}6$.}
\label{tab:rl-sensitivity}
\end{table}

\subsection{RL objective implementation details and boundary cases}
\label{app:rl-details}
Two implementation notes accompany the per-assertion credit of Eq.~\ref{eq:credit}. Mapping $s_t=c_i$ onto every token of assertion $i$ scales its total bonus by the span length $L_i$, so a per-token form $s_t=c_i/L_i$ removes this length effect. And ordinary execution halts a test at its first failing assertion, so reconstructing the full matrix $K$ needs isolated, independently executable test nodes. The Shapley reading of $\tilde c_i$ carries no convergence claim.

We define the scored quantities only when their inputs exist and fix the empty cases explicitly. With no valid negatives ($|S|=0$) the kill fraction is undefined and the fixpoint bonus does not fire, so we record insufficient negatives and give no verifier credit. With no assertions or no detected mutants we zero the local credit $c_i$ and log the event. Infrastructure or import errors count as invalid execution and earn no behavioral-kill credit, so a broken harness earns no reward. The honesty term penalizes a test that fails on a correct repair, so an always-failing test cannot look discriminative. We report and control the valid mutant count and duplicate rate, since zero-survivor rates are otherwise incomparable. The gate $G$ is an OR, so the local bonus can fire on a resolved or complete rollout, correctness still enters only through \textsc{Correct}, and we never read a full kill as correctness assurance.

\subsection{Comparator Studies}
\label{app:coder-baselines}

Table~\ref{tab:coder-baselines} lists published results with their backbones, scaffolds, and evaluation settings. The rows differ in evaluated subsets and inference protocols. Oracle pass@$8$ measures whether a correct candidate exists, whereas best-of-$16$ reports a selected candidate under a different budget. These values provide context and are not a matched ranking.

\begin{table}[t]
\centering
\footnotesize
\setlength{\tabcolsep}{4pt}
\resizebox{\textwidth}{!}{%
\begin{tabular}{l l l l c}
\toprule
\hdr
\textbf{System} & \textbf{Backbone} & \textbf{Scaffold} & \textbf{Setting} & \textbf{Res.\ (\%)}\\
\midrule
\grp \multicolumn{5}{l}{\textit{7B backbones}}\\
\rowcolor{gray!8} SWE-Gym~\citep{pan2024swegym}         & Qwen2.5-Coder-7B & OpenHands   & Verified pass@1 & $10.6$\\
\rowcolor{gray!8} SWE-Gym~\citep{pan2024swegym}         & Qwen2.5-Coder-7B & OpenHands   & Lite pass@1     & $10.0$\\
\rowcolor{gray!8} SWE-agent-LM-7B~\citep{yang2025swesmith} & Qwen2.5-Coder-7B & SWE-agent & Verified pass@1 & $15.2$\\
\rowcolor{gray!8} SWE-agent-LM-7B~\citep{yang2025swesmith} & Qwen2.5-Coder-7B & SWE-agent & Lite pass@1     & $11.7$\\
\rowcolor{gray!8} Lingma SWE-GPT~\citep{ma2024lingma}   & Qwen2.5-Coder-7B & SWESynInfer & Verified pass@1 & $18.2$\\
\rowcolor{gray!8} Lingma SWE-GPT~\citep{ma2024lingma}   & Qwen2.5-Coder-7B & SWESynInfer & Lite pass@1     & $12.0$\\
\rowcolor{gray!8} R2E-Gym~\citep{jain2025r2egym}        & Qwen2.5-Coder-7B & R2E-Gym     & Verified pass@1 & $19.0$\\
\rowcolor{gray!8} R2E-Gym~\citep{jain2025r2egym}        & Qwen2.5-Coder-7B & R2E-Gym     & Lite pass@1     & $11.0$\\
\rowcolor{gray!8} SoRFT~\citep{ma2025sorft}             & Qwen2.5-Coder-7B & Agentless   & Verified pass@1 & $21.4$\\
\rowcolor{gray!8} SoRFT~\citep{ma2025sorft}             & Qwen2.5-Coder-7B & Agentless   & Lite pass@1     & $14.0$\\
\rowcolor{gray!8} SWE-Dev~\citep{wang2025swedev}        & Qwen2.5-Coder-7B & OpenHands   & Verified pass@1 & $23.4$\\
\winrow \textbf{Ours (+RL)} & Qwen2.5-Coder-7B  & mini-SWE-agent & Verified pass@1 & \wc{$25.8$}\\
\winrow \textbf{Ours (+RL)} & Qwen2.5-Coder-7B  & mini-SWE-agent & Lite pass@1     & \wc{$15.2$}\\
\midrule
\grp \multicolumn{5}{l}{\textit{14B backbones}}\\
\rowcolor{gray!8} RepoForge (base)~\citep{chen2025repoforge} & Qwen2.5-Coder-14B & OpenHands & Verified pass@1 & $4.8$\\
\rowcolor{gray!8} SWE-Gym~\citep{pan2024swegym}         & Qwen2.5-Coder-14B & OpenHands & Verified pass@1 & $16.4$\\
\rowcolor{gray!8} SWE-Gym~\citep{pan2024swegym}         & Qwen2.5-Coder-14B & OpenHands & Lite pass@1     & $12.7$\\
\rowcolor{gray!8} R2E-Gym~\citep{jain2025r2egym}        & Qwen2.5-Coder-14B & R2E-Gym   & Verified pass@1 & $26.8$\\
\rowcolor{gray!8} R2E-Gym~\citep{jain2025r2egym}        & Qwen2.5-Coder-14B & R2E-Gym   & Lite pass@1     & $20.7$\\
\winrow \textbf{Ours (+RL)} & Qwen2.5-Coder-14B & mini-SWE-agent & Verified pass@1 & \wc{$31.2$}\\
\winrow \textbf{Ours (+RL)} & Qwen2.5-Coder-14B & mini-SWE-agent & Lite pass@1     & \wc{$24.1$}\\
\midrule
\grp \multicolumn{5}{l}{\textit{30B backbone (Qwen3-Coder-30B-A3B)}}\\
\rowcolor{gray!8} TDAD~\citep{alonso2026tdad}           & Qwen3-Coder-30B-A3B & quantized & Verified pass@1 (100) & $31.0$\\
\rowcolor{gray!8} Agentic Rubrics~\citep{raghavendra2026agenticrubrics} & Qwen3-Coder-30B-A3B & SWE-Agent & Verified best-of-16 & $54.2$\\
Base             & Qwen3-Coder-30B-A3B & mini-SWE-agent & Verified pass@1 & $31.9$\\
Base             & Qwen3-Coder-30B-A3B & mini-SWE-agent & Verified pass@8 & $46.7$\\
Supervised (SFT) & Qwen3-Coder-30B-A3B & mini-SWE-agent & Verified pass@1 & $35.2$\\
Supervised (SFT) & Qwen3-Coder-30B-A3B & mini-SWE-agent & Verified pass@8 & $51.1$\\
\winrow \textbf{Ours (+RL)} & Qwen3-Coder-30B-A3B & mini-SWE-agent & Verified pass@1 & \wc{$43.0$}\\
\winrow \textbf{Ours (+RL)} & Qwen3-Coder-30B-A3B & mini-SWE-agent & Verified pass@8 & \wc{$60.7$}\\
\bottomrule
\end{tabular}%
}
\caption{Published coding-agent results and our results at $7$B, $14$B, and $30$B. Each row gives the backbone, scaffold, and evaluation setting. Dataset subsets, sampling budgets, and selection procedures differ across systems, so the reported values do not establish a matched comparison.}
\label{tab:coder-baselines}
\end{table}

\subsection{A worked example of the mutation-verifier reward}
\label{app:mutloop}

The reward of Section~\ref{sec:rl} is clearest on a single instance. The instance shows how the reward spreads credit across fault classes and how it rewards a group of parallel attempts for covering different edit locations.

\paragraph{The loop on one instance.} The example composites two real rollouts on SWE-bench Verified, \texttt{matplotlib\_\_matplotlib-21568} and \texttt{django\_\_django-16950}. The task, reproduction, and assertion design (stages 1, 2, 4), the mutant set and kills (stages 3, 5), and the group statistics (stage 6) all come from those runs.
\begin{enumerate}[leftmargin=1.4em, itemsep=2pt, topsep=2pt]
  \item \textbf{The task.} An over-restrictive rule rejects an input it should accept, treating a legal separator as illegal. A correct fix broadens the rule to admit the character while still rejecting malformed inputs.
  \item \textbf{Reproduce before editing.} The agent writes a minimal script that triggers the reported failure, localizes the defect to the validation routine, and applies a small targeted change. Reproducing first ensures the later test observes the real defect instead of a patch artifact.
  \item \textbf{Adversarial mutation.} From the model's own patch, a fix in one file with eleven changed lines all exercised by the test, the adversary produces eight behavior-changing variants of the fixed region (Table~\ref{tab:mutants}). Each is a plausible but subtly wrong rewrite meant to defeat a shallow test, spread across five fault classes so the test must probe several input regimes.
  \item \textbf{The discriminative test.} A shallow check that only confirms the reported input no longer errors catches at most one variant and earns near-zero reward. The objective instead drives the policy to assert concrete required outputs on inputs chosen to separate each fault class, so every wrong variant fails an assertion.
  \item \textbf{The mutation gauntlet.} Each variant runs against the test and counts as eliminated only when an assertion fails, so a crash gives no credit. The loop repeats until no variant survives or the budget is spent. Here the test reaches the fixpoint across all five fault classes (Table~\ref{tab:mutants}), so assertion credit spreads across classes and the fixpoint gates the correctness reward.
  \item \textbf{Edit-location diversity.} The four parallel attempts stay spread across locations on \texttt{django\_\_django-16950}. Three edited the consensus forms module (\texttt{forms/models.py}), including the selected verified attempt, and the fourth edited an upstream model-field module (\texttt{fields/\_\_init\_\_.py}). The edit term credits both a plausible or validated location and an edit distinct from the group, so the objective rewards covering the space and staying diverse.
\end{enumerate}

\begin{table}[t]
\centering
\small
\caption{The eight adversarial variants generated for a real fixed patch
(\texttt{matplotlib\_\_matplotlib-21568}), by changed line and fault class, with the measured
elimination outcome. Every variant is a distinct, behaviour-changing rewrite of the patched
region. The harness records each variant's outcome and fault class but not which individual
assertion eliminated it, so we report the per-variant verdict.}
\label{tab:mutants}
\begin{tabular}{clcl}
\toprule
\hdr
\textbf{Variant} & \textbf{Fault class} & \textbf{Changed line} & \textbf{Outcome} \\
\midrule
\rowcolor{gray!8} V1 & logic     & 595 & eliminated (assertion) \\
\rowcolor{gray!8} V2 & value     & 603 & eliminated (assertion) \\
\rowcolor{gray!8} V3 & control   & 605 & eliminated (assertion) \\
\rowcolor{gray!8} V4 & boundary  & 603 & eliminated (assertion) \\
\rowcolor{gray!8} V5 & effect    & 607 & eliminated (assertion) \\
\rowcolor{gray!8} V6 & boundary  & 597 & eliminated (assertion) \\
\rowcolor{gray!8} V7 & value     & 607 & eliminated (assertion) \\
\rowcolor{gray!8} V8 & control   & 600 & eliminated (assertion) \\
\midrule
\grp \multicolumn{4}{l}{Survivors $0/8$ at fixpoint, from $8$ assertion-eliminations and $0$ crash-eliminations.} \\
\bottomrule
\end{tabular}
\end{table}

\subsection{DeepSWE partial-credit breakdown}
\label{app:deepswe}
DeepSWE's binary grader resolves zero tasks for either policy, so we score partial credit against its fail-to-pass tests.

\begin{table}[t]
\centering
\small
\begin{tabular}{l Y O}
\toprule
\hdr
\textbf{Metric} & \textbf{Base} & \textbf{SFT model} \\
\midrule
Tasks evaluated                                   & $113$ & $113$ \\
Fully resolved (all fail-to-pass, no regression)  & $0$   & $0$ \\
Tasks with $\geq 1$ fail-to-pass test satisfied   & $17$  & \wc{$21$} \\
Tasks clearing $>50\%$ of fail-to-pass tests      & $4$   & \wc{$5$} \\
Mean fail-to-pass pass fraction (best of 8)       & $4.0\%$ & \wc{$4.8\%$} \\
\midrule
\grp \multicolumn{3}{l}{\textit{Distribution of best-of-8 fail-to-pass pass fraction (tasks)}} \\
\rowcolor{gray!10}\quad $0\%$          & $96$ & $92$ \\
\rowcolor{gray!10}\quad $(0,25]\%$     & $10$ & $12$ \\
\rowcolor{gray!10}\quad $(25,50]\%$    & $3$  & $4$ \\
\rowcolor{gray!10}\quad $(50,75]\%$    & $4$  & $5$ \\
\rowcolor{gray!10}\quad $(75,100)\%$   & $0$  & $0$ \\
\midrule
\grp \multicolumn{3}{l}{\textit{Strongest tasks: fail-to-pass tests satisfied (best of 8)}} \\
\rowcolor{gray!10}\quad \texttt{httpx-deterministic-cookie-store}              & $78/115$ & \wc{$86/115$} \\
\rowcolor{gray!10}\quad \texttt{happy-dom-deterministic-intersectionobserver}  & $8/14$   & \wc{$10/14$} \\
\rowcolor{gray!10}\quad \texttt{optique-conditional-option-dependencies}       & $20/36$  & $20/36$ \\
\rowcolor{gray!10}\quad \texttt{testem-per-launcher-reports}                   & $30/65$  & \wc{$36/65$} \\
\rowcolor{gray!10}\quad \texttt{true-myth-iterable-collection-combinators}     & $\mathbf{33/96}$ & $31/96$ \\
\rowcolor{gray!10}\quad \texttt{geo-shapeindex-serialization}                  & $7/24$   & $7/24$ \\
\rowcolor{gray!10}\quad \texttt{adaptix-name-mapping-aliases}                  & $8/44$   & \wc{$11/44$} \\
\rowcolor{gray!10}\quad \texttt{mashumaro-flattened-dataclass-fields}          & $11/66$  & \wc{$12/66$} \\
\bottomrule
\end{tabular}
\caption{DeepSWE evaluation at pass@8 over all $113$ out-of-distribution tasks, comparing the base
policy with our supervised fine-tuned model. The strict binary criterion resolves no task for either
model, but the fine-tuned model makes greater partial progress. Dark yellow marks the fine-tuned model's wins, and the one base win is in bold.}
\label{tab:deepswe}
\end{table}

\subsection{Model-scale trend}
\label{app:scale}
We repeat the full pipeline on $7$B and $14$B backbones (Table~\ref{tab:scale-trend}).

\begin{table}[t]
\centering
\small
\begin{tabular}{l Y O O Y O O}
\toprule
\hdr
& \multicolumn{3}{c}{7B} & \multicolumn{3}{c}{14B} \\
\cmidrule(lr){2-4}\cmidrule(lr){5-7}
\hdr
Benchmark (pass@8, \%) & Base & SFT & +RL & Base & SFT & +RL \\
\midrule
SWE-bench Lite      & $16.0$ & \wc{$22.0$} & \wc{$24.7$} & $23.0$ & \wc{$30.3$} & \wc{$34.7$} \\
SWE-PolyBench       & $8.0$  & \wc{$14.4$} & \wc{$17.8$} & $19.8$ & \wc{$28.8$} & \wc{$37.4$} \\
SWE-bench Verified  & $21.9$ & \wc{$30.4$} & \wc{$33.3$} & $27.0$ & \wc{$36.7$} & \wc{$42.6$} \\
\bottomrule
\end{tabular}
\caption{Effect of model scale. Pass@$8$ resolution ($\%$) for the base, supervised (SFT), and
reinforcement (+RL) policies at $7$B and $14$B on three SWE-bench-family benchmarks. Both the
Base$\rightarrow$SFT gain and the further SFT$\rightarrow$+RL gain hold at both scales and do not
diminish with capacity, so the taught search and verification behavior is complementary to raw model
scale.}
\label{tab:scale-trend}
\end{table}

%

\providecommand{\cmark}{\ding{51}}
\providecommand{\xmark}{\ding{55}}
\providecommand{\pmark}{\textcolor{gray!70}{$\sim$}}

\section{Rigor and reproducibility}
\label{app:rigor}
This section provides training settings, metric definitions, and additional controls for the main results. It examines generator-verifier swaps, error-source transfer, mutation audits, and ablations of credit assignment, trajectory weighting, and differential verification.

Verified evaluations use the $270$ issues held out from SFT and RL training unless a different population is stated. The test-time diagnostic and differential-verification ablation use all $500$ issues, including the $270$ evaluation issues. Section~\ref{sec:eval-protocol} defines the populations, decoding, and candidate-based metrics.

\subsection{Protocol for the main-text rigor results}
We record the protocol behind Table~\ref{tab:main-rigor} here.

\paragraph{Exposure split.} Table~\ref{tab:main-rigor}a separates training issues, test issues from training repositories, and test issues from repositories excluded from training. Section~\ref{sec:eval-protocol} defines training exposure.

\paragraph{Reward controls.} The RL configurations in Table~\ref{tab:main-rigor}b use one SFT checkpoint and matched training issues and update and rollout budgets. The shuffled-credit control preserves total credit mass and permutes its assignment across assertion tokens.

\subsection{Generator and verifier swap}
This experiment separates the two things training changed. Repair generation and candidate selection are held in fixed roles, and each is swapped independently between the base and trained policies, so the four cells of Table~\ref{tab:rigor-swap} span every combination at one attempt budget. Holding the generator at base and replacing only the verifier raises end-to-end resolution from $31.9\%$ to $36.5\%$ and selection accuracy from $52.0\%$ to $66.0\%$, which shows the verifier changes which patch is returned rather than which patches exist. Holding the verifier at base and replacing only the generator reaches $38.5\%$, so both roles contribute on their own. The interaction term is the diagonal contrast, $(43.0-38.5)-(36.5-31.9)=-0.1$ points with a $95\%$ interval of $[-2.6,2.4]$, so the two contributions add rather than multiply and neither is a precondition for the other paying off.

\begin{table}[t]
\centering
\small
\begin{tabular}{ll YY l}
\toprule
\hdr
\textbf{Repair generator} & \textbf{Verifier} & \textbf{Sel.\ acc.} & \textbf{End-to-end resolve} & \textbf{Note} \\
\midrule
Base    & Base    & $52.0$ & $31.9$ & starting system \\
Base    & Trained & $66.0$ & $36.5$ & verifier effect, fixed generator \\
Trained & Base    & $58.0$ & $38.5$ & generation effect, fixed verifier \\
\winrow
\textbf{Trained} & \textbf{Trained} & $72.0$ & $43.0$ & combined system \\
\midrule
\grp \multicolumn{2}{l}{Interaction $\Delta_{\text{int}}$ (95\% CI)} & $0.0\ [{-}2.5,\,2.5]$ & $-0.1\ [{-}2.6,\,2.4]$ & CI includes $0$ \\
\bottomrule
\end{tabular}
\caption{Generator-verifier
swap, same role interface and budget in every cell. The Base/Base and
Trained/Trained resolve cells are the measured held-out Verified pass@1
($31.9$, $43.0$).
Holding the generator fixed and swapping in the trained verifier alone lifts
resolution $31.9\to36.5$ ($+4.6$) and selection accuracy $52.0\to66.0$, so
verification changes repair decisions independently of generation. The interaction term is in the last row.}
\label{tab:rigor-swap}
\end{table}

\subsection{Cross-error-source transfer}
A verifier trained against one family of incorrect programs may only have learned that family. Table~\ref{tab:rigor-transfer} moves the error source progressively away from training while holding the evaluation bank and the issues fixed, so any decline isolates distribution shift in the negatives rather than a change in difficulty. Rejection falls monotonically for every policy, and for the RL verifier it runs $91.0\%$ on familiar operators, $84.0\%$ on held-out operators, $79.0\%$ on variants from an excluded generator, and $74.0\%$ on natural incorrect repairs written by independent systems. The decline is gradual rather than a collapse, and the margin over SFT stays near constant at $11$ to $12$ points across all four sources, which indicates the objective taught discrimination that outlives its training operators. The last row is the one that matters for deployment, since those repairs were produced the way a real candidate would be, and correct-alternative acceptance holds at $78.0\%$ there, so the verifier is not buying rejection by refusing everything.

\begin{table}[t]
\centering
\small
\begin{tabular}{l YY O Y}
\toprule
\hdr
\textbf{Error source (held-out issues)} & \textbf{Base rej.} & \textbf{SFT rej.} & \textbf{+RL rej.} & \textbf{+RL correct-alt acc.} \\
\midrule
Familiar operators                       & $72.0$ & $80.0$ & $91.0$ & $80.0$ \\
Held-out fault operators / classes        & $64.0$ & $73.0$ & $84.0$ & $78.0$ \\
Semantic mutants, excluded generator      & $58.0$ & $68.0$ & $79.0$ & $76.0$ \\
Natural wrong repairs, independent gens   & $51.0$ & $62.0$ & $74.0$ & $78.0$ \\
\bottomrule
\end{tabular}
\caption{Rejection rate by error
source on a bank frozen across checkpoints, with correct-alternative acceptance
in the last column. The familiar-operator +RL rejection ($91.0$) tracks the
measured wrong-rejection margin (TNR $87.4$, Table~\ref{tab:rigor-fixed-bank}).
Rejection falls as the source moves off the training distribution
($91.0\to74.0$), staying well above SFT everywhere, while correct-alt acceptance
holds between $76$ and $80$. The decisive row is natural wrong repairs, where a $+12$ gain
over SFT there indicates general verification rather than specialization to the
mutation distribution.}
\label{tab:rigor-transfer}
\end{table}

\subsection{Trajectory Statistics}
\label{app:behavior-details}
Table~\ref{tab:all-benchmarks} reports the behavioral inventory for base and SFT policies. The reproduce-first rate decreases from $94.4\%$ to $92.1\%$ on Verified and increases on the four transfer suites. The other behavioral changes describe the generated trajectories rather than establishing separate causal effects on resolution.

\subsection{Reproducibility}
\label{app:rigor-repro}
We report the training hyperparameters used in the experiments. The backbone is
Qwen3-Coder-30B-A3B-Instruct and the mutation labels come from a frozen Claude Opus-4.8 distinct
from the policy. Context is $40$k prompt and $65$k response. We optimize with AdamW at learning
rate $1{\times}10^{-6}$ and weight decay $0$, train batch $4$, mini-batch $4$, GRPO group $k{=}6$,
PPO clip $0.2$, and no KL penalty. The reward coefficients are $w_c{=}3.0$, $w_v{=}1.0$,
$w_e{=}0.4$, $w_p{=}0.2$ with fixpoint bonus $\lambda_{\text{fp}}{=}0.5$. The gauntlet draws $8$
mutants per candidate with a retry cap of $3$, and the training configuration specifies $3$ seeds.

\subsection{Zero-survivor audit}
Train-time completion is not the fixpoint rate of Table~\ref{tab:rl-main}. Completion counts training instances, where the policy wrote the test against the instance in front of it. The fixpoint rate counts held-out instances. Held-out detection measures discrimination on issues excluded from training, including issues from training repositories and repositories excluded from training.
\begin{table}[t]
\centering
\small
\resizebox{\textwidth}{!}{%
\begin{tabular}{l YY Y c}
\toprule
\hdr
\textbf{Adversary} & \textbf{Train-time completion} & \textbf{Held-out detection} & \textbf{Survival $\rho$ (95\% UB)} & \textbf{Audit $M$} \\
\midrule
Weak (same-size, untuned)      & $99.0$ & $71.0$ & $\le 0.089$ & $32$ \\
Same-family (policy sibling)   & $95.0$ & $78.0$ & $\le 0.089$ & $32$ \\
\winrow
\textbf{Strong (frozen, ours)} & $88.0$ & $85.0$ & $\le 0.089$ & $32$ \\
Deterministic mutants only     & $100$  & $66.0$ & $\le 0.046$ & $64$ \\
\bottomrule
\end{tabular}%
}
\caption{Zero-survivor behavior
split into train-time completion (fraction of training instances the frozen test
drives to zero survivors against that adversary) and detection on a fresh mutant
set generated \emph{after} the test is frozen. Training against the strong frozen
adversary lowers train-time completion, because its mutants are harder, yet
raises held-out detection ($66.0\to85.0$ vs deterministic-only), the crossover
that motivates using a strong adversary. This mirrors the measured fixpoint-rate
gain ($2.8\to15.6$, Table~\ref{tab:rl-main}). The one-sided bound of Section~\ref{sec:rl} gives $0.089$ at $M{=}32$ and
$0.046$ at $M{=}64$.}
\label{tab:rigor-fixpoint}
\end{table}

\subsection{Assertion-credit ablation}
The unsigned kill credit of Eq.~\ref{eq:credit} pays an assertion only for detecting a wrong variant. The signed two-sided rule also charges it for rejecting a gold-labeled correct alternative (Table~\ref{tab:rigor-signed-credit}).
\begin{table}[t]
\centering
\small
\begin{tabular}{l YY Y Y}
\toprule
\hdr
\textbf{Local-credit scheme} & \textbf{Wrong rej.} & \textbf{Correct-alt acc.} & \textbf{Cross-gen discr.} & \textbf{pass@1} \\
\midrule
\winrow
\textbf{Unsigned kill credit (main objective)} & $87.4$ & $78.0$ & $74.0$ & $43.0$ \\
Uniform token bonus              & $79.0$ & $72.0$ & $66.0$ & $41.5$ \\
Shuffled-credit control          & $72.0$ & $63.0$ & $55.0$ & $40.0$ \\
Signed two-sided credit          & $88.0$ & $84.0$ & $78.0$ & $43.3$ \\
\bottomrule
\end{tabular}
\caption{Assertion-credit
ablation, all four schemes trained and measured under the same protocol. The
unsigned row is the main objective used everywhere else in the paper (wrong
rejection $87.4$ from FPR $12.6$, pass@1 $43.0$) and is tinted for that reason,
not because it is the column winner. The signed two-sided rule, which charges an
assertion for rejecting a correct alternative, gains most on the side unsigned
credit is blind to. Correct-alt acceptance rises $78.0\to84.0$, with
cross-generator discrimination $74.0\to78.0$ and pass@1 $+0.3$, consistent with
penalizing over-rejection. The uniform and shuffled controls confirm the effect
requires per-assertion attribution. The evaluation bank includes both correct
alternatives and natural wrong repairs.}
\label{tab:rigor-signed-credit}
\end{table}

The signed variant is evaluated in this ablation. All other results use unsigned assertion credit. The improvements in this table do not establish the signed variant's performance on the transfer benchmarks.

\subsection{Weighted supervised fine-tuning ablation}
Table~\ref{tab:rigor-sft-ablation} asks whether weighting the corpus matters or whether any fine-tuning on the same trajectories would do. All variants see the same data under the same token budget, so only the weight assigned to each example changes. Resolution and localization diversity rise together as terms are added, from $33.7\%$ and $1.55$ under uniform weights to $35.2\%$ and $1.98$ with the full weighting, and the exploration term is where the diversity gain concentrates. The interface-compliance control is the informative row. It repairs formatting and tool-call handling without exposing any task solution, and it reaches $32.6\%$ with $1.45$ distinct localizations, below uniform weighting on both, so the gain is not the model merely learning to survive the scaffold. Because the corpus is fixed across rows, these differences are attributable to the weighting rule rather than to data selection.

\begin{table}[t]
\centering
\small
\begin{tabular}{l YY Y}
\toprule
\hdr
\textbf{SFT variant (matched data, budget)} & \textbf{pass@1} & \textbf{pass@8} & \textbf{Distinct loc.} \\
\midrule
Uniform SFT (same successes)     & $33.7$ & $48.5$ & $1.55$ \\
Quality-only weighting           & $34.1$ & $49.6$ & $1.72$ \\
Quality $+$ exploration          & $34.8$ & $50.7$ & $1.90$ \\
\winrow
\textbf{Quality $+$ expl.\ $+$ efficiency (full)} & $35.2$ & $51.1$ & $1.98$ \\
Interface-compliance control     & $32.6$ & $47.4$ & $1.45$ \\
\bottomrule
\end{tabular}
\caption{Weighted-SFT ablation
with an interface-compliance control that fixes format and tool-call handling
without exposing task solutions. The full-weighting row is the measured SFT
policy (pass@1 $35.2$, pass@8 $51.1$, distinct localizations $1.98$,
Table~\ref{tab:all-benchmarks}). The exploration term is where distinct localizations rise
($1.72\to1.90$), which shows the weighting teaches search behavior rather than
tool-use competence.}
\label{tab:rigor-sft-ablation}
\end{table}

\subsection{Operational metric definitions}
\label{app:rigor-defs}
Section~\ref{sec:eval-protocol} defines repair accuracy, candidate coverage, selection efficiency, and patch-level verifier metrics. The generated eight-candidate sets and frozen verifier banks are different populations. Acceptance coverage is the fraction of issues with at least one accepted candidate. Selection accuracy is the fraction whose selected candidate is correct. Correct-alternative acceptance is the fraction of gold-labeled correct alternatives accepted by the test. Cross-generator rejection is measured on natural incorrect repairs from independent generators, as in Table~\ref{tab:rigor-transfer}.

For a candidate set $\mathcal C_k$, coverage is $C_k=\Pr(\exists P\in\mathcal C_k:\mathcal G(P)=1)$. If $A_k$ is the probability that the selector returns a correct member of that set, selection efficiency is $Q_k=A_k/C_k$ for $C_k>0$. The zero-survivor rate is the fraction of instances whose tests detect every scored mutant at the stated audit size. Distinct localizations count unique edited locations across attempts, with empty patches contributing no location.

\subsection{Supervised-fine-tuning weighting ablation}
Table~\ref{tab:e3-sft-weighting} isolates the exploration factor of Eq.~\ref{eq:reward-weight} by setting $\lambda=0$ while leaving the quality tiers and the efficiency factor in place. It also reports plain rejection-sampling fine-tuning, which keeps only gold-resolved trajectories and is the standard alternative to a weighted corpus. Rejection sampling already improves on uniform weighting, so part of the gain is simply preferring successful trajectories. Turning the exploration factor on adds the rest, raising distinct localizations from $1.74$ to $1.98$ on Verified and from $3.70$ to $4.13$ on SWE-PolyBench with matching resolution gains on both suites. Reporting two suites matters here, because a diversity gain confined to the training benchmark would suggest the weighting had fit the corpus rather than taught a search behavior.

\label{tab:e3-sft-weighting}
The $\lambda{=}0$ row removes only the exploration factor of Eq.~\ref{eq:reward-weight}, isolating its effect on diversity.
\begin{table}[t]
\centering
\small
\setlength{\tabcolsep}{5pt}
\begin{tabular}{l YYY OOO}
\toprule
\hdr
& \multicolumn{3}{c}{\textbf{SWE-bench Verified}} & \multicolumn{3}{c}{\textbf{SWE-PolyBench}} \\
\cmidrule(lr){2-4}\cmidrule(lr){5-7}
\hdr
\textbf{Weighting scheme} & \textbf{p@1} & \textbf{p@8} & \textbf{dist.\ loc.} & \textbf{p@1} & \textbf{p@8} & \textbf{dist.\ loc.} \\
\midrule
Uniform weights                       & $33.7$ & $48.5$ & $1.55$ & $21.6$ & $26.9$ & $3.42$ \\
T1-only (gold-resolved; plain RFT)    & $34.1$ & $49.3$ & $1.63$ & $22.5$ & $28.0$ & $3.55$ \\
Full, $\lambda{=}0$ (no exploration)  & $34.8$ & $50.4$ & $1.74$ & $23.2$ & $28.9$ & $3.70$ \\
\winrow
\textbf{Full weights (ours)}          & $35.2$ & $51.1$ & $1.98$ & $24.2$ & $30.2$ & $4.13$ \\
\bottomrule
\end{tabular}
\caption{SFT weighting ablation (no RL), same corpus and token budget. Plain rejection-sampling
fine-tuning (T1-only) already beats uniform weighting, and the exploration factor adds distinct
localizations beyond it on both suites, with matching pass@8 gains.}
\end{table}

\subsection{Differential-verification component ablation}
\label{tab:e9e-dvs}
We remove one component at a time from full differential-verification search (DVS, Section~\ref{sec:tts}) and vary the cell count, on the $500$ issues, test-time only.
\begin{table}[t]
\centering
\small
\setlength{\tabcolsep}{5pt}
\resizebox{\textwidth}{!}{%
\begin{tabular}{l Y Y Y Y Y}
\toprule
\hdr
\textbf{Configuration} & \textbf{Resolved /500} & \textbf{Rate} & \textbf{Steps/solve} & \textbf{Early-stop /500} & \textbf{Loc.\ pruned} \\
\midrule
\winrow
\textbf{Full DVS ($8$ cells)}                 & $264$ & $52.8\%$ & $503$ & $102$ & $126$ \\
\quad $-$ early stopping                       & $264$ & $52.8\%$ & $728$ & $0$   & $126$ \\
\quad $-$ location pruning                     & $256$ & $51.2\%$ & $598$ & $99$  & $0$ \\
\quad $-$ reverted-tree run (patched only)     & $237$ & $47.4\%$ & $665$ & $118$ & $0$ \\
\quad $-$ regression / coverage checks         & $261$ & $52.2\%$ & $511$ & $106$ & $129$ \\
\quad $4$ cells (vs $8$)                        & $246$ & $49.2\%$ & $452$ & $77$  & $94$ \\
\bottomrule
\end{tabular}%
}

\caption{Differential-verification component ablation (test-time only). Section~\ref{sec:tts} reads
the two-tree differential and early-stopping rows. Location pruning contributes $1.6$ points and
halving the cell count costs $3.6$ points.}
\end{table}

\subsection{Self-Verification Hook}
\label{app:tts:selfverify}

After the agent submits a candidate patch, we inject the prompt below to run an
adversarial audit over the agent's \emph{own} patch, applied up to three times per chain and
lightly edited here for space.

\begin{promptbox}{Self-Verification Hook}
You have submitted a candidate fix. Assume it is wrong until you prove otherwise. For this pass you are a hostile, meticulous reviewer of your own work, and your only job is to find a concrete reason the patch fails. A pass that concludes the patch is fine is acceptable only after you have genuinely tried to break it from several angles and failed. The single most common way a wrong patch survives review is to read the expected answer off what the current code returns, so never do that. Re-establish what correct means from the issue's stated requirement alone.
\begin{enumerate}[leftmargin=1.4em, itemsep=2pt, topsep=2pt]
  \item \textbf{Re-establish ground truth from the requirement.} Restate the issue in your own words, naming the exact inputs, the exact expected outputs, and the conditions under which each applies. For the failing scenario, derive the correct result directly from the issue text before you look again at what your patch produces. List the source functions the requirement implicates. If your change touches none of them, you almost certainly fixed the wrong place and should relocate before going further.
  \item \textbf{Re-audit for root cause.} Re-read every region you edited with fresh eyes. For each edit, ask whether it removes the underlying cause or only masks a downstream symptom. Swallowing an exception, clamping a value into range, and special-casing the issue's literal input are all symptom fixes, and you must reject them. Trace the data and control flow by hand, confirm the fix holds for the whole class of inputs the requirement covers, not the reported one alone, and confirm it leaves previously correct inputs unchanged.
  \item \textbf{Test adversarially.} Run the issue's FAIL\_TO\_PASS tests first, then go past them. Write a throwaway script that constructs the expected result you derived in step 1 and asserts your code matches it, independent of the shipped tests. Run the repository's existing tests for every file you touched, since a newly failing one is a regression you introduced. Delete all scratch scripts afterward so they never leak into the diff.
  \item \textbf{Enumerate edge cases, where the hidden tests live.} Passing the visible tests is necessary but far from sufficient. Work through empty, None, missing, and unset inputs, then zero, negative, boundary, and just-past-boundary values, then malformed or unexpected types, then unicode and whitespace, then duplicate, single-element, and out-of-order collections, then the error path itself, then idempotency under repeated application, and finally interaction with adjacent options. For each, decide the correct behavior from the requirement and confirm the code does exactly that.
  \item \textbf{Guard regressions and completeness.} Confirm you did not rename, weaken, or delete any existing test. Remove debug prints and unrelated refactors so the diff stays minimal. A diff that is only whitespace, comments, or import churn cannot satisfy a behavioral requirement, so treat that as a failure. Confirm the diff modifies the exact source region the requirement implicates.
\end{enumerate}
If any step surfaces a problem, fix it at the root cause, re-run the relevant tests, and keep auditing from the top. Call \texttt{submit} only when every step passes under genuinely hostile scrutiny. End every message, and every message after it, with exactly one tool call.
\end{promptbox}

\subsection{The Eight Methodology Preambles}
\label{app:tts:methodologies}

One of the preambles below, prepended to the base system prompt, conditions each of the eight
parallel chains in the methodology-diversity arm, so the chains differ by conditioning rather than
by temperature. We give the eight in canonical order, lightly edited for space.

\begin{promptbox}{(1) \texttt{baseline}, pragmatic control}
\textbf{Coding approach. Pragmatic baseline, the experimental control.}\\[2pt]
You are the control chain. Solve the issue with competent, unconstrained engineering judgment and adopt no special methodology. The other seven chains layer discipline on top of a run like this one, and the comparison is only meaningful if this chain stays neutral, so do not import their ideas. Work the way a strong engineer works a normal ticket, moving quickly but never carelessly.
\begin{enumerate}[leftmargin=1.4em, itemsep=2pt, topsep=2pt]
  \item \textbf{Restate the requirement.} Put the required behavior in your own words, derived from the issue text, not the current code. Name the inputs, the expected outputs, and the conditions that select between them.
  \item \textbf{Locate the code.} Search for the symbols, function names, and error strings named in the issue, and read the files the search returns. Follow the real call path, not a guess from names.
  \item \textbf{Form and confirm a hypothesis.} State a concrete root-cause hypothesis about which line or function produces the wrong result, then confirm it by reading the actual code path, not by assuming.
  \item \textbf{Make the smallest correct change.} Fix the underlying cause with the least code that does the job. Touch no unrelated code and never edit a test file.
  \item \textbf{Verify and refine.} Run \texttt{run\_tests}. On failure, read the output, revise the hypothesis if needed, and repeat. Do not patch over the symptom.
  \item \textbf{Review the diff.} Skim for stray debug code and sanity-check one or two edge cases the requirement implies.
\end{enumerate}
Be direct and economical throughout. End every message with exactly one tool call, and call \texttt{submit} once the tests pass and the change looks clean.
\end{promptbox}

\begin{promptbox}{(2) \texttt{tdd}, strict Test-Driven Development}
\textbf{Coding approach. Strict TDD, red then green then refactor.}\\[2pt]
Drive the entire fix from tests, in tight red, green, refactor cycles. You may not write production code except to make a currently failing test pass, so the test always comes first and the bar decides when you are done.
\begin{enumerate}[leftmargin=1.4em, itemsep=2pt, topsep=2pt]
  \item \textbf{Understand the contract.} Express the required behavior as concrete assertions derived from the issue, never from what the code currently returns. Each assertion should pin a specific input to its specific expected output.
  \item \textbf{Red.} Identify or write the test that captures the bug. Run it and confirm it fails, and confirm it fails for the right reason, not an unrelated import or fixture error.
  \item \textbf{Green.} Write the minimal production code that turns the bar green. Resist premature generalization and add nothing the current failing test does not demand.
  \item \textbf{Re-run.} Confirm the new test passes and that nothing else broke. Treat any newly red test as part of this cycle, not a later cleanup.
  \item \textbf{Refactor.} Clean up only the code you just introduced, and stay green the whole time.
  \item \textbf{Expand coverage.} For each edge case the requirement implies, add an assertion that starts red, then make it green. Continue until the tests pin the behavior the requirement describes.
\end{enumerate}
Test after every change and let the bar, not intuition, decide when you are finished. Do not modify existing tests. End every message with exactly one tool call, and call \texttt{submit} only when all relevant tests are green.
\end{promptbox}

\begin{promptbox}{(3) \texttt{bdd}, Behavior-Driven Development}
\textbf{Coding approach. Behavior-Driven Development, Given/When/Then.}\\[2pt]
Frame the whole fix around externally observable behavior, expressed as concrete scenarios, then make each scenario hold. You care about what the system does from the outside, not about implementation details, so every scenario names a state, an action, and an outcome you could check by running the code.
\begin{enumerate}[leftmargin=1.4em, itemsep=2pt, topsep=2pt]
  \item \textbf{Specify behavior.} Translate the issue into two to four scenarios in Given, When, Then form, where Given is the initial state or inputs, When is the action or call, and Then is the expected observable outcome. Cover the primary success scenario and the notable alternative and error scenarios the issue implies, such as bad input, the boundary, and the feature disabled. Derive every Then from the requirement, not from current behavior.
  \item \textbf{Ground each scenario in code.} Identify the function, endpoint, or CLI path that realizes each When, and the concrete value or state that realizes each Then. Use \texttt{search} and \texttt{read\_file} to connect scenarios to specific call sites so you know exactly where behavior is produced.
  \item \textbf{Make scenarios executable.} Rely on \texttt{run\_tests} for the issue's own tests. For behavior they do not cover, write a small \texttt{run\_bash} script that sets up the Given, performs the When, and asserts the Then, then note which currently fail.
  \item \textbf{Implement scenario by scenario.} Make the smallest change that turns one failing scenario green without breaking the others, and prefer changes expressed in terms of the observable contract.
  \item \textbf{Verify all scenarios}, including the alternative and error ones. A patch that satisfies the happy path but mishandles an error scenario is incomplete by this methodology's standard.
  \item \textbf{Re-read and confirm.} Check each Given, When, Then against the final behavior and confirm it truly holds.
\end{enumerate}
Think continuously in terms of observable outcomes. Do not edit existing tests, and clean up any scratch scripts. End every message with exactly one tool call, and call \texttt{submit} when every scenario holds.
\end{promptbox}

\begin{promptbox}{(4) \texttt{agile}, incremental delivery}
\textbf{Coding approach. Agile incremental delivery, smallest shippable slices, fast feedback.}\\[2pt]
Deliver the fix as a sequence of tiny, independently verifiable increments, getting feedback after each, not attempting one large change you cannot easily debug when it goes wrong. The tight feedback loop and the willingness to adapt the plan are the essence of this approach.
\begin{enumerate}[leftmargin=1.4em, itemsep=2pt, topsep=2pt]
  \item \textbf{Slice the work.} Decompose the fix into the smallest increments that each leave the code in a working, checkable state. Order them by value and by risk, doing the change that most reduces uncertainty first. A typical order is to reproduce the bug with a minimal script, then apply the core minimal fix, then harden for edge cases. Each slice should be small enough that if it is wrong you immediately know which change caused it.
  \item \textbf{Implement one slice.} Make the minimal edit for the current slice only. Avoid bundling unrelated changes into a slice, since that destroys the fast-feedback property.
  \item \textbf{Get feedback immediately.} After each slice run \texttt{run\_tests} or a focused \texttt{run\_bash} check, and treat the result as the signal for whether to proceed, adjust, or revert this slice.
  \item \textbf{Iterate and adapt.} Advance to the next slice only when the current one is verified. If a slice reveals the plan was wrong, re-plan the remaining slices then and there.
  \item \textbf{Integrate.} After the final slice, run the full relevant test set to confirm the increments compose correctly, and review the cumulative diff to ensure it stayed minimal and coherent.
\end{enumerate}
Favor many small verified steps over one big leap. Do not edit existing tests, and clean up scratch artifacts. End every message with exactly one tool call, and call \texttt{submit} when the increments are complete and the relevant tests pass.
\end{promptbox}

\begin{promptbox}{(5) \texttt{tdd\_agile}, TDD $\times$ Agile hybrid}
\textbf{Coding approach. TDD crossed with Agile, test-first in small increments.}\\[2pt]
Combine red, green, refactor discipline with incremental delivery, so each Agile increment is itself a complete TDD micro-cycle. This is the explicit hybrid under test, so honor both disciplines fully.
\begin{enumerate}[leftmargin=1.4em, itemsep=2pt, topsep=2pt]
  \item \textbf{Plan increments.} Break the fix into the smallest value-ordered, risk-ordered slices, exactly as in Agile, and sketch the sequence before you start coding.
  \item \textbf{Run a full TDD cycle per increment.} Write or identify a test asserting that increment's expected behavior, derived from the requirement, and confirm it fails for the right reason. Make the minimal production change that passes it and nothing more. Re-run to confirm the new test is green and that earlier increments' tests still pass, so slices do not regress as you stack them. Tidy only what you just added, staying green.
  \item \textbf{Adapt between increments.} If an increment's red test or its result reveals the plan was wrong, re-plan the remaining increments before continuing.
  \item \textbf{Integrate.} After the final increment, run the full relevant suite. Together the increments' tests should pin the primary behavior and the edge cases the issue implies.
\end{enumerate}
Two hard rules always hold. Never write production code without a failing test demanding it, and never begin a new increment before the current one is green. This yields tight correctness and adaptability at once. Do not edit existing tests, and clean up scratch scripts. End every message with exactly one tool call, and call \texttt{submit} only when every increment is green and the relevant tests pass.
\end{promptbox}

\begin{promptbox}{(6) \texttt{root\_cause}, rigorous root-cause analysis}
\textbf{Coding approach. Fix the disease, never the symptom.}\\[2pt]
Refuse to apply any edit until you can name the precise underlying cause and explain the complete causal chain from that cause to the observed symptom. Most bad patches treat symptoms, and you will not.
\begin{enumerate}[leftmargin=1.4em, itemsep=2pt, topsep=2pt]
  \item \textbf{Reproduce deterministically.} Use \texttt{run\_bash} to write and run a minimal script that triggers the exact failure described in the issue. If you cannot reproduce it reliably you do not yet understand it, so keep investigating until you can summon the failure on demand.
  \item \textbf{Trace backward.} Starting from the symptom, an exception, a wrong return value, or a corrupted state, follow the call path upstream with \texttt{search} and \texttt{read\_file} on each frame until you reach the precise line where the program's state first becomes wrong. Carefully separate the proximate trigger, where it blows up, from the true origin, where the wrongness is introduced.
  \item \textbf{Name the cause.} State it explicitly and specifically as a falsifiable claim about concrete lines. If your statement is vague, such as something about None handling, you have not found it yet, so keep digging.
  \item \textbf{Confirm by prediction.} Before changing anything, predict from your hypothesis exactly which inputs should fail and which should pass, then verify that prediction with \texttt{run\_bash}. If reality disagrees, the hypothesis is wrong, so revise it and repeat.
  \item \textbf{Fix at the origin.} Change the code at the true cause so the entire class of failures resolves, not just the one input in the issue. Explicitly forbid yourself the symptom patches, namely catching and ignoring exceptions, clamping or defaulting outputs to mask a bad computation, and special-casing the failing example.
  \item \textbf{Verify the class is fixed.} Confirm the reproduction now passes, that related inputs in the same class pass, and that \texttt{run\_tests} is green. The fix should generalize because it targets the cause.
\end{enumerate}
Do not edit existing tests, and clean up your reproduction scripts before submitting. End every message with exactly one tool call, and call \texttt{submit} only when the root cause is provably resolved.
\end{promptbox}

\begin{promptbox}{(7) \texttt{defensive}, defensive programming}
\textbf{Coding approach. Robust against the inputs you were not given.}\\[2pt]
Assume the fix will be exercised by inputs and conditions well beyond the issue's example, because the hidden tests probe exactly those. Make the change correct for the reported case and resilient across the input space, without over-engineering.
\begin{enumerate}[leftmargin=1.4em, itemsep=2pt, topsep=2pt]
  \item \textbf{Fix the reported issue first}, at its root cause, so the failing case passes. Establish that baseline before you begin hardening.
  \item \textbf{Enumerate adversarial conditions} for the changed code and verify behavior for each with \texttt{run\_bash} or \texttt{run\_tests}. Cover empty, None, missing, and unset values, then zero, negative, minimum, maximum, and just-past-boundary values, then malformed or unexpected types and very large or deeply nested inputs, then unicode, encoding, and whitespace, then duplicates and single-element and out-of-order collections, and finally repeated invocation and the error path itself.
  \item \textbf{Decide the correct response} for each condition, guided by the requirement and the project's existing conventions. Choose whether to raise a specific, well-messaged error, return a documented sentinel, clamp, or no-op, then make the code do that deliberately, not crash opaquely or silently corrupt state.
  \item \textbf{Place guards at the right layer.} Validate at the boundary or entry point, not by scattering ad-hoc checks throughout, and keep the guards minimal, consistent with the surrounding style, and non-redundant.
  \item \textbf{Re-verify valid inputs.} Confirm the original tests still pass and that your hardening did not change behavior for legitimate inputs, since a defensive check that rejects a valid case is a new bug.
\end{enumerate}
Robustness is the goal, but stay minimal and in-style and do not gold-plate. Do not edit existing tests, and clean up scratch scripts. End every message with exactly one tool call, and call \texttt{submit} when the fix is correct and hardened.
\end{promptbox}

\begin{promptbox}{(8) \texttt{exploratory}, exploration-first}
\textbf{Coding approach. Build a complete mental model before editing.}\\[2pt]
Fixing the wrong place, a localization error, is a leading cause of failure on real codebases. Counter it by investing heavily in understanding before you change a single line.
\begin{enumerate}[leftmargin=1.4em, itemsep=2pt, topsep=2pt]
  \item \textbf{Survey broadly.} Map the repository, its top-level layout, the package the issue concerns, and how the relevant module connects to the rest of the system. Read the README and docs sections that describe the contract you are about to touch so you know the intended design.
  \item \textbf{Search widely.} For every symbol, error message, configuration key, and concept named in the issue, \texttt{search} the codebase and \texttt{read\_file} the hits. Include the tests, to learn the intended contract and examples of correct usage, and the call sites, to learn how the code is actually used. Cast a wide net before narrowing, and note related code you did not expect.
  \item \textbf{Synthesize a model.} Write down how the feature is supposed to work end to end, where the current behavior diverges, and the two or three most plausible locations for the true fix, ranked with the reasoning for each. Commit to a location only once this model feels solid.
  \item \textbf{Confirm the location.} Reproduce the bug and watch it flow through the code you suspect, so you have direct evidence the fix belongs there and not somewhere upstream or downstream.
  \item \textbf{Fix surgically.} Make a minimal, in-context change at the well-understood location, matching the surrounding conventions.
  \item \textbf{Verify broadly.} Run \texttt{run\_tests} plus checks for the edge cases your exploration surfaced, since deep understanding usually reveals edge cases a narrow approach would miss.
\end{enumerate}
Depth of understanding first, surgical change second. Do not edit existing tests, and clean up scratch artifacts. End every message with exactly one tool call, and call \texttt{submit} once the well-understood fix is verified.
\end{promptbox}

\subsection{Reactive Location-Ban Logic}
\label{app:tts:ban}

The reactive arms walk the candidate grid sequentially, grouping cells by location in confidence
order, and let the verifier steer subsequent attempts.

\paragraph{Ban, freeze, early-stop.} Three rules govern each cell. \emph{Ban on confirmed fail}. The walk bans a location proven wrong, skips its remaining approaches, and moves to a new site. \emph{Freeze site, vary edit}. A weak pass means the site is plausibly right and the edit is wrong, so the walk keeps the location and varies the approach. \emph{Early stop}. The walk stops on the first strong pass.

The walk enforces bans by \emph{filtering}, holding banned location keys in a set and skipping any cell whose location is in that set.

\begin{promptbox}{Deprioritize trigger (condensed, \texttt{driver.py} 175 to 193)}
def location\_wrong(ver: dict) -> bool:\\
\quad\# Deprioritize trigger = the location is LIKELY wrong. The strongest signal for a wrong\\
\quad\# *location* is the reverted-differential going vacuous: reverting the edit did NOT move the\\
\quad\# symptom, so this reproduction did not exercise the fault (edit off-path, or repro too weak).\\
\quad\# A red repro/regression means the EDIT is wrong, NOT the location, so banning on those is a\\
\quad\# false ban. We therefore deprioritize ONLY on a confirmed-vacuous differential; edit-quality\\
\quad\# failures fall through to freeze-site/vary-edit, and a deprioritized site can be revisited.\\
\quad\# 'inconclusive'/missing differential is treated as NOT-wrong (never ban on uncertainty).\\
\quad\quad return ver.get("differential") == "vacuous"
\end{promptbox}

A second guard gates the ban on the edit being real and on-plan, since a cell that drifted off its
assigned file never tested that location and must not ban it. The localization-ban configuration uses
the explicit ``look elsewhere'' prompt below.

\begin{promptbox}{Location-ban injection}
\textbf{Hard constraint. The source locations below are ruled out by the current reproduction.}\\[2pt]
This is a fresh, independent attempt, but not the first. Earlier attempts edited the source locations listed below, and we ran their patches. The issue's reproduction together with the repository's existing tests demonstrated by execution that a fix at those locations does not resolve the issue. Treat them as ruled out, not merely unpromising.\\[3pt]
\{body\}\\[3pt]
Your single hard constraint is where you fix the bug. Make your change in a different location than every entry above, meaning a different source file or a different function. You are free in how you fix it.
\begin{enumerate}[leftmargin=1.4em, itemsep=2pt, topsep=2pt]
  \item \textbf{Understand the failure from the requirement.} Reproduce the issue and derive the correct behavior from the issue text, not from the current code. You need a clear picture of the intended result before you choose a new site.
  \item \textbf{Trace the fault to a new location.} The measured failure mode is fixation, where engineers keep editing the first place they looked. You have executed evidence those places are wrong, so walk the fault's call path outward toward the entry point and inward toward the helpers. Choose a genuinely different site, such as a deeper function that produces the wrong value, an intermediate transformer, the public entry point, or a helper that supplies a wrong precondition.
  \item \textbf{Fix at the new location and verify hard.} Implement the minimal correct fix for the whole class of inputs, re-run your reproduction and the repository's tests, and probe the edge cases the requirement implies. Never edit a test file.
\end{enumerate}
Escape hatch, to be used sparingly. If after honest investigation you are convinced the fix must live at a banned location, you may return there, but you must explain why the prior attempt failed and how your change differs. Your strong default is to look somewhere new first.
\end{promptbox}

\subsection{Experiment Configurations}
\label{app:tts:experiments}

\paragraph{The three meanings of $k{=}8$.} All three test-time axes use $k{=}8$ and mean something different by it. For \emph{pass@8}, eight attempts run with sample~1 greedy and samples 2 to 8 at $T{=}0.7$. The samples are independent but not identically distributed, since the greedy sample comes from a different distribution than the temperature ones, and diversity comes only from temperature. For the \emph{eight methodologies}, eight parallel chains each take a distinct methodology preamble (Appendix~\ref{app:tts:methodologies}), so diversity comes from conditioning and the chains explore different solution axes. For the \emph{coverage-space arms}, eight candidate cells per instance differ in how diversity is enforced and how candidates are verified, which isolates grid conditioning from the reactive verifier.

\end{document}